%% file: paper.tex
\documentclass[conference]{IEEEtran}
\IEEEoverridecommandlockouts
\usepackage{cite}
\usepackage{amsmath,amssymb,amsfonts}
\usepackage{algorithmic}
\usepackage{graphicx}
\usepackage{textcomp}

\usepackage{enumitem}

\usepackage{multicol}
\usepackage{caption}
\usepackage{xspace}
\usepackage{multirow}
\usepackage{booktabs}
\usepackage{bm}
\usepackage{xcolor}
\def\BibTeX{{\rm B\kern-.05em{\sc i\kern-.025em b}\kern-.08em
    T\kern-.1667em\lower.7ex\hbox{E}\kern-.125emX}}

\begin{document}

\newtheorem{definition}{Definition}
\newcommand{\paratitle}[1]{\vspace{1.5ex}\noindent\textbf{#1}}
\newcommand{\ie}{\emph{i.e.,}\xspace}
\newcommand{\aka}{\emph{a.k.a.,}\xspace}
\newcommand{\eg}{\emph{e.g.,}\xspace}
\newcommand{\etal}{\emph{et al.}\xspace}
\newcommand{\wrt}{\emph{w.r.t.}\xspace}

\newcommand{\model}{GARCIA}

\title{{\model}: Powering Representations of Long-tail Query with Multi--granularity Contrastive Learning}

\author{\IEEEauthorblockN{Weifan Wang$^\star$,
Binbin Hu$^\star$,
Zhicheng Peng$^\star$,
Mingjie Zhong,
Zhiqiang Zhang,
\\
Zhongyi Liu,
Guannan Zhang,
Jun Zhou\IEEEauthorrefmark{2}
}
\IEEEauthorblockA{Ant Group, China }
\IEEEauthorblockA{\IEEEauthorrefmark{0}\{weifan.wwf, bin.hbb, zhicheng.pzc, mingjie.zmj, lingyao.zzq, zhongyi.lzy, zgn138592, jun.zhoujun\}@antgroup.com} 
\thanks{$^\star$Contribute equally.} \thanks{\IEEEauthorrefmark{2}Corresponding author.}
}

\maketitle

\begin{abstract}
Recently, the growth of service platforms brings great convenience to both users and merchants, where the service search engine plays a vital role in improving the user experience by quickly obtaining desirable results via textual queries.
Unfortunately, users' uncontrollable search customs usually bring vast amounts of long-tail queries, which severely threaten the capability of search models. 
Inspired by recently emerging graph neural networks (GNNs) and contrastive learning (CL), several efforts have been made in alleviating the long-tail issue and achieve considerable performance.
Nevertheless, they still face a few major weaknesses. Most importantly, they do not explicitly utilize the contextual structure between heads and tails for effective knowledge transfer, and intention-level information is commonly ignored for more generalized representations.

To this end, we develop a novel framework {\model}, which exploits the graph based knowledge
transfer and intention based representation generalization in a contrastive setting. In particular, we employ an adaptive encoder to produce informative representations for queries
and services, as well as hierarchical structure aware representations of intentions.
To fully understand tail queries and services, we equip {\model} with a novel multi-granularity contrastive learning module, which powers representations through knowledge transfer, structure enhancement and intention generalization.
Subsequently, the complete {\model} is well trained in a pre-training\&fine-tuning manner.
At last, we conduct extensive experiments on both offline and online environments, which demonstrates the superior capability of {\model} in improving tail queries and overall performance in service search scenarios.

\end{abstract}


\section{Introduction}
\begin{figure}
  \includegraphics[width=0.48\textwidth]{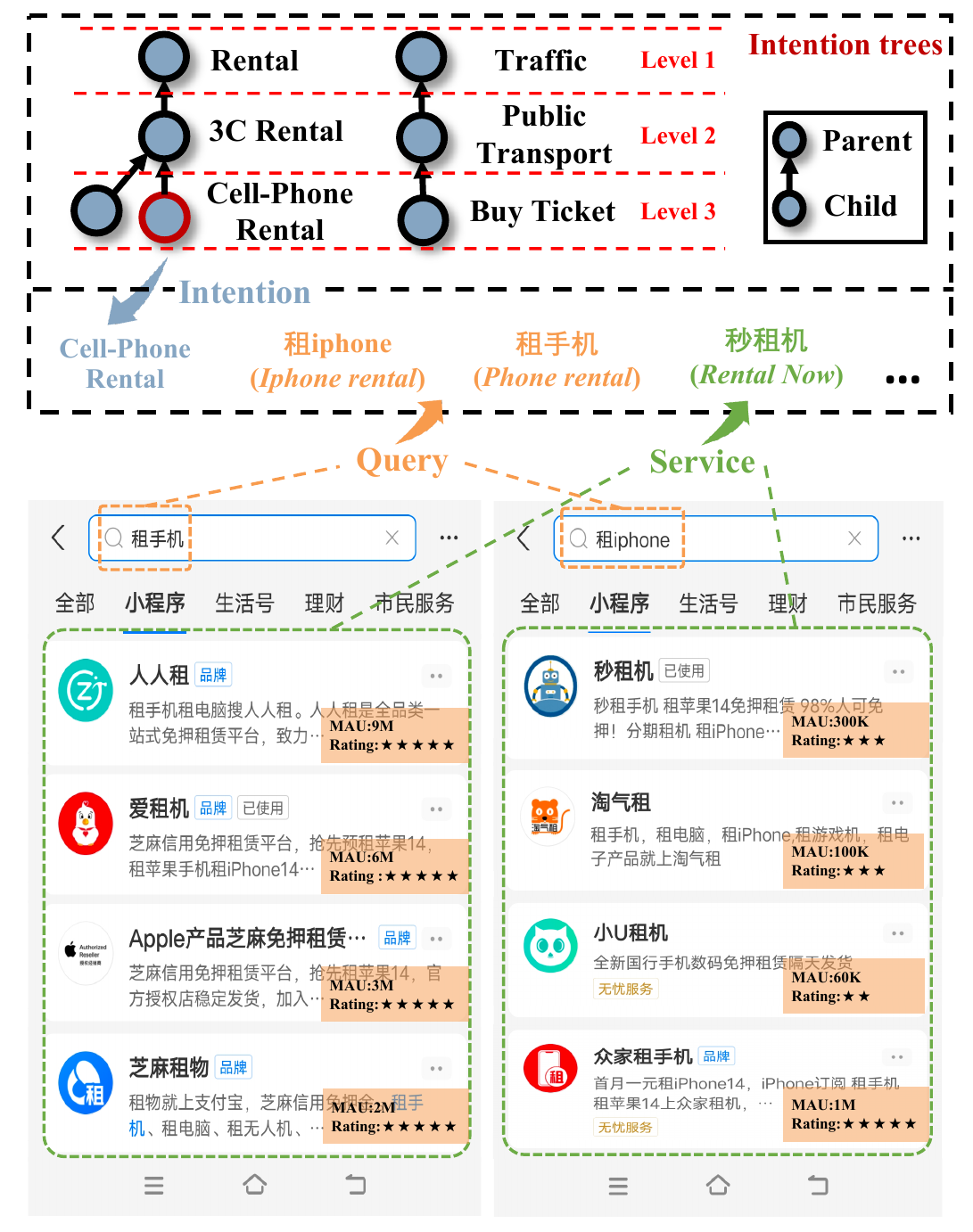}
  \caption{An example of service search scenario with intention trees in Alipay. We mark the quality of each service via MAU and Authoritative Rating (``Rating'' for short in the figure) at its bottom-right corner. A larger MAU and Rating means a higher quality of the service. Best viewed in color.}
  \label{fig:example}
\end{figure}

In recent years, convenience promote the rapid growth of service platforms (\eg Alipay and WeChat), which allow merchants and users to provide and obtain services through operating small programs, respectively. As an indispensable component, service search aims at delivering desired results that meet the information needs of customers from a huge collection of services~\cite{liu2021que2search}. However, building the service search engine in real-world applications is challenging, suffering from the notorious long-tail issue in user behaviors, which means that few high-frequency queries (\ie head queries) dominate in
search input while low-frequency queries (\ie tail queries) have a
very low probability of occurrences~\cite{zhu2022enhanced}. Take the Alipay service search as an example, the uncontrollable search customs of users inevitably bring vast amounts of long-tail queries. As shown in Fig.~\ref{fig:example},  the intention ``Cell-Phone Rental" easily induces two similar queries ``Phone Rental'' and ``Iphone Rental'', while the latter is a long-tail query, leading to unsatisfying results with low MAU and authoritative ratings~\footnote{The definitions of the MAU and the authoritative rating are detailed in the experiment part.}. Through comprehensive statistics, we are surprised to find that only 1\% of the queries account for more than 90\% of the traffic in the service search scenario  of Alipay. Since surfacing an undesirable list of services given the user textual query greatly harms the user experience, it becomes critical to address the long-tail issue to improve the overall quality of the ranking list of services under textual queries.

A plethora of  research works have been devoted to  tackling this problem in the field of recommendation and search, earlier studies including dataset-resampling~\cite{he2009learning} and loss-reweighting~\cite{cui2019class} strategies. Unfortunately, the success of these heuristic methods highly hinges on domain knowledge about the control of sampling rate and weight of head/tail queries in the loss function, leading to sub-optimal and unstable performance. On the other hand, recent years have seen a surge of efforts in exploring the long-tail issue in the view of representation learning, which roughly fall into two lines. The first line introduces the recently emerging graph neural networks to flexibly incorporate various side-information (\ie social~\cite{fan2019graph,kou2020efficient,wang2022intent}, heterogeneous~\cite{hu2018leveraging,li2020transn,niu2020dual} and knowledge graphs~\cite{wang2019kgat,wang2020semantic,yang2021newslink}) for context enrichment, especially for tails. Inspired by the flourishing of contrastive learning (CL), another line investigates into enhancing representations of tails through auxiliary contrastive supervisions with stochastic augmentations~\cite{wei2021contrastive}. 

Although these methods have achieved performance to some extent, they still suffer serious limitations.
First, these methods~\cite{wang2022intent,yang2021newslink,li2020transn,niu2020dual} mainly focus on the characterization of tails based on well-establish graphs or augmentation strategies, where the contextual bridge between heads and tails (\eg belonging to the same brand) is not explicitly utilized for knowledge transfer.
Second, users’ behaviors in service search scenarios are usually intention oriented, which is not considered in current methods. As shown in Fig.~\ref{fig:example}, head query ``Phone Rental'' and tail query ``Iphone Rental'' could be expressed by the same intention ``Cell-Phone Rental'', indicating such intention-level context have the potential ability for generalizing the representations of tail queries and services.
Third,  CL augmented GNN methods~\cite{wei2021contrastive,wu2021self,yu2022graph} typically improve the representations of tails via performing node- and graph-level augmentations in random, which may be vulnerable to noises of  graph structure in real-world scenarios.

To break loose from the above limitations, given the real-world service search scenario, we exploit the graph based knowledge transfer and intention based representation generalization in a contrastive setting, with the aims of improving the performance of tail queries, thus yielding the promising improvement of overall performance.  However, its solution is non-trivial, given two key challenges:
\begin{itemize}[leftmargin=*]
    \item \emph{C1: Effective distillation and transfer of useful knowledge. } Generally, the long-tail issue causes the large distribution gap between head and tail queries, and structural patterns of tail queries are usually ignored due to their minority. Hence, it is imperative to design a new paradigm that effectively distills and transfers knowledge from heads to tails based on adaptive encoding and local structural aware enhancement.
    \item \emph{C2: Flexible characterization and incorporation of intention-level context.} In real-world scenarios, intention-level information is generally organized through various relations and hierarchical dependencies, forming intention trees, as shown at the top of Fig.~\ref{fig:example}. Such a tree-like structure brings a  good generalized ability for learning queries and services, which are commonly overlooked in current works. Thus, it is of crucial importance to flexibly characterize and incorporate the intention-level context with the careful consideration of  hierarchical structure.
\end{itemize}

With these challenges in mind, we propose \textbf{\model}, a novel \underline{G}r\underline{A}ph based service sea\underline{R}ch framework with multi-granularity \underline{C}ontrast\underline{I}ve le\underline{A}rning. 
As the basis of {\model}, we elaborately establish a service search graph consisting of interaction and correlation relations between queries and services. 
Subsequently, we employ two individual GNNs to encode head and tail queries/services atop of the service search graph respectively (\emph{C1}), while  a bottom-up aggregation strategy is also designed to capture the  hierarchical structure of the intention tree  (\emph{C2}).
We take inspiration from the recent developments of contrastive learning~\cite{zhu2020deep,wei2021contrastive,xie2022contrastive,luo2022dualgraph}, and thus equip {\model} with a multi-granularity contrastive learning module that comprehensively understands tail queries/services through three aspects:
i) transferring knowledge from the head to help the learning of the tail, 
ii) enhancing representations via local structure, and 
iii) generalizing representations based on the intention tree (\emph{C1} \& \emph{C2}).
At last, we pre-training the {\model} in a multi-task manner, followed by a fine-tuning procedure tailored for our service search scenario.
It is noteworthy that {\model} performs CL via explicitly exploiting various relations in the service search scenario, endowing {\model} potential robustness against noise in real-world applications.  
The contributions of this work are summarized as follows:
\begin{itemize}[leftmargin=*]
    \item \textbf{General Aspects:} We emphasize the crucial importance of knowledge transfer and intention generalization for improving the quality of query/service representations in graph based service search.  
    \item \textbf{Novel Approach:} We propose {\model} that not only adaptively encodes tails and heads and flexibly captures hierarchical structure among intentions, but also is equipped with multi-granularity contrastive learning module that powers representation through knowledge transfer, structure enhancement and intention generalization.
    \item \textbf{Multifaceted Experiments and Outstanding Performance:} Extensive experiments on three offline datasets reveal that {\model} brings solid performance improvement across all metrics (\ie AUC, GAUC and NDCG@K) for both tail and head queries. Moreover, {\model} is carefully implemented and deployed in the online environment and achieves significant performance. Both of observations demonstrate the superior capability of GARCIA tailored for the long-tail issue in the service search scenarios.

\end{itemize}

\input{sec-rel}

\input{sec-pre}

\section{The Proposed Approach}\label{sec:model}
In this section, to center on the long-tail issue  in real-world service search scenarios, we present {\model}, a novel \underline{G}r\underline{A}ph based service sea\underline{R}ch framework with multi-granularity \underline{C}ontrast\underline{I}ve le\underline{A}rning. Generally, the main procedure of {\model} has three main parts:
\begin{itemize}[leftmargin=*]
    \item \textbf{Adaptive encoding for queries/services and intentions.} {\model} employs a GNN atop of the service search graph to produce informative representations for queries and services, while a bottom-up aggregation strategy is also designed to capture various relations and hierarchical dependencies derived from the intention tree.
    \item \textbf{Multi-granularity contrastive learning.} With the aims of fully understanding tail queries, {\model} leverages contrastive learning for representation powering through knowledge transfer, structure enhancement and intention generalization.
    \item \textbf{Model learning in a pre-training\&fine-tuning manner.} The proposed {\model} is first pre-trained through multi-granularity contrastive loss, and the pre-trained parameters are utilized to initialize the parameters,  followed by the training tailored for our service search scenarios.
\end{itemize}

An overview of {\model} is shown in Fig.~\ref{fig:framework}. In the ensuing parts, we will zoom into each well-designed component of {\model}.


\subsection{Adaptive Encoding}
The natural graph (tree) structures upon queries/services (intentions) allow messages to pass over the nodes with local contextual information, which offers the potential opportunities to produce powerful representations with the aggregation schema. 
Hence, we employ GNNs to perform effective representation learning for queries/services on the service search graph, while intentions are encoded through a bottom-up aggregation way to model inherent characteristics of intentions in different levels.

\subsubsection{Encoding queries/services on the service search graph \label{sec:graph_encode}}
To effectively inject GNNs into the service search scenarios for facilitating representation learning, the core is to define the pipeline of ``aggregation'', ``update'' and ``readout'' (Eq. 1). Formally, the overall encoding process of {\model} could be summarized as follows:
\begin{equation}
    \label{eq:adap-enc-1}
    \begin{split}
    \text{Aggregate:} \ \ &    \bm{m}_q^{(l)} = \text{Tanh}(\mathbf{W}^{\mathcal{A}}\sum_{v \in \mathcal{N}_q}{\alpha_{q,v} [\bm{z}_v^{(l)}||\bm{e}_{q,v}]}), \\
     \text{Update:} \ \ &    \bm{z}_q^{(l+1)} = \text{ReLU}(\mathbf{W}^{\mathcal{U}}[\bm{z}_q^{(l)}||\bm{m}_q^{l}]),\\
     \text{Readout:} \ \ &    \bm{z}_q = \frac{1}{L+1}\sum_{l=0}^{L}{\bm{z}_q^{(l)}},\\
     \end{split}
\end{equation}
where $\mathbf{W}^\mathcal{A}$ and  $\mathbf{W}^\mathcal{U}$ are the weight matrices for ``aggregate'' and ``update'' step, respectively, $\bm{e}_{q,v}$ denotes the attributes on edges (\eg click-through rate and various correlations), and $\alpha_{q,v}$ controls the decay factor on the aggregation \wrt edge ``$q \leftarrow v$'', implemented by the recent emerging attention mechanism~\cite{velivckovic2017graph}. After $L$-layers aggregation and updating, we combine the representations of all layers for the final representations. The representations of services can be
obtained analogously, denoted as $\bm{z}_s$

Recalling that the distribution of feedback in the service search scenario is very skewed, head and tail queries/services in the graph may exhibit various properties and thus require diverse models~\footnote{The split strategy of head and tail queries/services is described in Sec.~\ref{sec:ktcl}.}. Since sharing a unified model may decrease the expressiveness of {\model}, we separately equip the head and tail queries/services with individual GNN encoders. In short, for a head query $q$ or service $s$, we represent it as $\bm{z}^{head}_q$ and $\bm{z}^{head}_s$, while  a tail query $q$ or service $s$ is encoded as $\bm{z}^{tail}_q$ or $\bm{z}^{tail}_s$. 
We argue that such an adaptive operation is necessary since a unified GNN is prone to ignore patterns of tail queries and services due to their minority.



\subsubsection{Encoding intention on the intention tree}
Intuitively, intention-level information could help the generalization of representations, which potentially avoids underlying noises. To effectively capture intention-level characteristics, the inherent relations and dependencies among intentions are necessary to be considered. Hence, we take full advantage of structural information by performing a bottom-up aggregation strategy upon our intention tree. Specifically, we learn the representation for target intention $i$ on intention tree $\mathcal{T}$ via aggregating its child nodes $v \in \mathcal{N}^{\mathcal{T}}_v$. Here, we adopt a GCN-like aggregator for $H$-levels intention tree aggregation.
\begin{equation}
    \label{eq:adap-enc-2}
    \bm{z}^{(h+1)}_i = \sigma(\mathbf{W}^{\mathcal{T}}(\bm{z}^{(h)}_i + \sum_{v \in \mathcal{N}^{\mathcal{T}}}{\bm{z}^{(h)}_v})),
\end{equation}
where $\mathbf{W}^{\mathcal{T}}$ is the weight matrix, and $\bm{z}^{(0)}_i$ is initialized by the learnable embedding table. Since the aggregation process is from leaf intentions to root intentions, the representations of intentions are level-related, \ie aware of hierarchical structure.
For the sake of brevity and readability, we omit the superscript and rewrite the representation of the intention $i$ as  $\bm{z}^{\mathcal{T}}_i$.

\subsection{Multi-granularity Contrastive Learning}
Until now, we have obtained the representations of queries and services enriched by our well-established service search graph. Unfortunately, purely aggregating neighbor information still fails to the full understanding of tail queries/services, since various underlying relations are commonly ignored.
Hence, we come up with integrating the multi-granularity contrastive learning tailored for the representation learning of tail queries/services, which mainly lies in three aspects:
i) transferring knowledge from the head to help the learning of the tail,
ii) enhancing representations via local structure, and
iii) generalizing representations based on the intention tree.

\subsubsection{Knowledge transfer oriented CL \label{sec:ktcl}}
Aiming at facilitating the representation learning of tail queries, an intuitive inspiration is to transfer credible knowledge from head queries to the tail. In the main service search scenario of Alipay, we are surprised to find that 1\% of the queries account for more than 90\% of the traffic, which discloses that 
i) head queries contain rich feedback information, and 
ii) tail queries are apt to be under-fitting due to their minority, making them unpredictable.
To tackle these issues, we propose the Knowledge Transfer oriented CL (KTCL) module to make similar head and tail queries reachable in the embedding space.

Firstly, we split the whole query set $\mathcal{Q}$ into head part $\mathcal{Q}^{head}$ and tail part $\mathcal{Q}^{tail}$ based on our domain knowledge: 
$\mathcal{Q}^{head}$  contains the top 10 thousand queries which occupy the most exposure during the past month 
and the remaining queries form $\mathcal{Q}^{tail}$. 
Obviously, a more delicate split strategy could promise better performance, and we leave it as future work. 
For each tail query $q^{tail} \in \mathcal{Q}^{tail}$, we are devoted to pick up a  head query $p^{head} \in \mathcal{Q}^{head}$ to consist an anchor pair $<q^{tail}, p^{head}>$ , satisfying following criteria:
\begin{itemize}
    \item The head query $p^{head}$ has the most semantic-level relevance with the tail query $q^{tail}$.
    \item The head query $p^{head}$ and the tail query $q^{tail}$ share the same correlations, \eg city, brand and category.
    \item Only the head query $p^{head}$ with the most exposure would be picked up.
\end{itemize} 
Based on the InfoNCE~\cite{he2020momentum}, we construct the KTCL loss for queries with in-batch negative sampling strategy as follows:
\begin{equation}
\label{eq:ktcl-Q}
    \mathcal{L}^{\mathcal{Q}}_{KTCL} = -\sum_{q^{tail} \in \mathcal{Q}^{tail}}\log \frac{\exp(\cos(z^{tail}_q, z^{head}_p) / \tau)}{\sum_{k \in \mathcal{B}^{head}}\exp(\cos(z^{tail}_q, z^{head}_k) / \tau)}, 
\end{equation}
where $\tau$ is the temperature hyper-parameter of softmax and $\mathcal{B}^{head}$ denotes the head queries in batch during the training procedure. 
On the service side, due to the query-level split, a service $s$ may be simultaneously encoded as $z^{tail}$ and $z^{tail}$, and thus the semantic alignment between them should be promised. Hence, we develop the following auxiliary contrastive loss in KTCL:
\begin{equation}
\label{eq:ktcl-S}
\begin{split}
     \mathcal{L}^{\mathcal{S}}_{KTCL} = &- \sum_{s \in \mathcal{S}} \left( \log \frac{\exp(\cos(\bm{z}^{head}_s, \bm{z}^{tail}_s)/ \tau) }{\sum_{s' \in \mathcal{B}} \exp(\cos(\bm{z}^{head}_s, \bm{z}^{tail}_{s'})/ \tau)}  \right.\\
     & \left. + \log \frac{\exp(\cos(\bm{z}^{tail}_s, \bm{z}^{head}_s)/ \tau) }{\sum_{s' \in \mathcal{B}} \exp(\cos(\bm{z}^{tail}_s, \bm{z}^{head}_{s'})/ \tau)}   \right).
\end{split}
\end{equation}
We put the above losses together for the complete KTCL objective function:
\begin{equation}
\label{eq:ktcl}
    \mathcal{L}_{KTCL}  = \mathcal{L}^{\mathcal{Q}}_{KTCL} + \mathcal{L}^{\mathcal{S}}_{KTCL}.
\end{equation}

In this way, KTCL encourages effective knowledge transfer from head queries to tail queries. As the foundation of such a transfer process, the quality of representation learning for tails and heads is of crucial importance, which drives the incorporation of structure enhancement and intention generalization.

\subsubsection{Structure enhancement oriented CL }
Inspired by the layer-by-layer aggregation mechanism of GNNs, we introduce the Structure Enhancement oriented CL (SECL) module for the representation powering with local structure. In particular, the SECL module regards the initial representations (\ie $\bm{z}^
{tail(0)}_q, \bm{z}_q^{head(0)}, \bm{z}_s^{head(0)}, \bm{z}_s^{tail (0)}$) as the anchor. The structure information involved in aggregation for learning query/service representation forms the positives, and the other representation in the current batch forms the negatives. Formally, the SECL loss for queries could be defined as follows~\footnote{For sake of  brevity, we omit superscript ``head'' and ``tail'' in SECL and IGCL module, since SECL and IGCL are applied both of head and tail queries/services}:
\begin{equation}
\label{eq:secl-Q}
\begin{split}
    \mathcal{L}_{SECL}^{\mathcal{Q} } & =  \\
    & -\frac{1}{L}\sum_{l = 1}^{L}  \sum_{q \in \mathcal{Q}}\log\frac{\exp(\cos(\bm{z}^{(l)}_q, \bm{z}^{(0)}_q) / \tau)}{\sum_{k \in  \mathcal{B}}\exp(\cos(\bm{z}^{(l)}_k, \bm{z}^{(0)}_q) / \tau)}.
\end{split}
\end{equation}
Analogously, we could formulate the SECL loss for services as $\mathcal{L}_{SECL}^{\mathcal{S}}$, and combine the two parts of losses as follows:
\begin{equation}
\label{eq:secl}
    \mathcal{L}_{SECL}  = \mathcal{L}^{\mathcal{Q}}_{SECL} + \mathcal{L}^{\mathcal{S}}_{SECL}.
\end{equation}
Overall, the SECL module highlights the effective exploitation of local structure, which have the following two-fold advantages:
i) Compared with the aggregation of GNNs, SECL supplements the explicit and coherent interaction of local neighbors, contributing to the final meaningful representation.
ii) SECL encourages a balanced information utilization in graphs, which potentially prevents the {\model} from uncontrollable noise derived from aggregation.

\subsubsection{Intention generalization oriented CL}
As mentioned above, users' behaviors in service search scenarios are usually intention oriented, which highlights the crucial importance of intention to the generalization of query/service representations. In previous works, most of the attempts directly utilize intentions as sparse features for prediction, which ignores the natural relations and hierarchical dependencies among intentions, leading to sub-optimal performance. Based on the intention representations encoded by the bottom-up aggregation strategy, we further explicitly characterize the hierarchical structure as well as the correlation between intention and queries/services in the view of the objective function, and thus propose the  Intention Generalization oriented CL (IGCL). In particular, given the intention tree $\mathcal{T}$, once a query $q$ is associated with an intention $i$, besides the intention $i$, we recursively collect its parents in $\mathcal{T}$, denotes as $\mathcal{P}_{q,i}$. 
After that, the proposed IGCL module is to minimize the following function:
\begin{equation}
\label{eq:igcl-Q}
    \begin{split}
     \mathcal{L}_{IGCL}^{\mathcal{Q}} & =  \\
   &-\sum_{q \in \mathcal{Q}} \frac{1}{|\mathcal{P}_{q, i}|}\sum_{j \in \mathcal{P}_{q, i}}\log \frac{\exp(\cos(\bm{z}_q, \bm{z}^{\mathcal{T}}_j)/ \tau) }{\sum_{k \in \mathcal{D}_{p, j}}\exp(\cos(\bm{z}_q, \bm{z}^{\mathcal{T}}_k) / \tau)},
    \end{split}
\end{equation}
where $\mathcal{D}_{p, j}$ is the negative set for IGCL, which mainly consists of two parts:
i) ``Hard'' negatives: the intentions in the same intention tree with the query $q$ and having the same level with intention $i$, and 
ii) ``Easy'' negatives: the intentions in another intention tree with an arbitrary query $q$ and having the same level with intention $i$.
The objective on the service side is identical, denoted as $\mathcal{L}_{IGCL}^{\mathcal{S}}$, and we combine them for the final IGCL objective function as:
\begin{equation}
\label{eq:igcl}
    \mathcal{L}_{IGCL} = \mathcal{L}_{IGCL}^{\mathcal{Q}} + \mathcal{L}_{IGCL}^{\mathcal{S}}.
\end{equation}
In this way, we explicitly incorporate the hierarchical structure among intentions into the contrastive learning, which alleviates the data sparsity and avoids underlying noises.

\subsection{Model Learning}
We follow the wide-adopted pre-training\&fine-tuning manner to train the {\model} for our service search scenario.
\subsubsection{Pre-training}
We first pre-train our model on the multi-granularity contrastive learning, which involves KTCL, SECL and IGCL. Hence, the overall pre-training objective for {\model} is summarized as follows:
\begin{equation}
    \mathcal{L}_{\mathcal{P}} = \mathcal{L}_{KTCL} + \alpha \mathcal{L}_{SECL} + \beta \mathcal{L}_{IGCL},
\end{equation}
where ${\alpha}$ and $\beta$  are both  hyper-parameters to balance the weights of the three modules in the pre-training stage.
\subsubsection{Fine-tuning}
Then we utilize the pre-trained parameters to initialize the parameters, and then adopt the service search task to train {\model} for online serving. Given a query $q$ and service $i$, the click probability is calculated by:
\begin{equation}
    \label{eq:pred}
    \hat{y}_{q, s} = \sigma(\text{MLP}([\bm{z}_q || \bm{z_s}])),
\end{equation}
where $\text{MLP} (\cdot)$ is a two-layer perceptron neural network with ReLU as the activation function.
And we adopt the widely used binary cross entropy as the final objective loss:
\begin{equation}
\label{eq:main-loss}
    \mathcal{L}_{\mathcal{F}} = -\sum_{<q,s>}y_{q,s}\log(\hat{y}_{q, s}) + (1 - y_{q, s})\log(1 - \hat{y}_{q, s}), 
\end{equation}
where $y_{q,s}$ denotes the ground truth. 

\section{Experiments \label{sec:exp}}
In this section, we perform multifaceted experiments to demonstrate the effectiveness of {\model} in offline and online environments. In summary, we aim at answering the following four research questions:
\textbf{RQ1}: Does the proposed {\model} outperform other state-of-the-art methods for the service search task?
\textbf{RQ2}: Does the proposed {\model} benefit from our well-designed components (\ie  multi-granularity contrastive learning and individual encoding for tails and heads)?
\textbf{RQ3}: How do the key hyper-parameters impact the performance of {\model}?
\textbf{RQ4}: How about the deployment of {\model} in the real-world service search scenario, and its performance in the online environment?



\subsection{Datasets}
\subsubsection{Industrial Datasets}
\label{sec-industrial}
We sample a real-world large-scale dataset from users' visit logs in Alipay, which contains the browse and click behaviors in September of 2022~\footnote{This dataset is only used for academic research, which has been encrypted and desensitized.}. In particular, the dataset includes more than $2.0 \times 10^7$ users and about $8.0 \times 10^5$ services with $3.89 \times 10^9$ interactions. 
In terms of search PV, we find that the cumulative search PV~\footnote{PV means Page View.} of the queries in the top 1\% accounts for about 90+\% of the overall search PV. Such a skewed distribution discloses that a large number of queries are not fully exposed, leading to the under-fitting issue and biased estimation of current models on these queries. 
To better understand the stable and robust performance of our proposed model, we chronologically split the whole data into three sub-dataset, namely \textbf{Sep. A} (2022/09/01 $\sim$ 2022/09/10), \textbf{Sep. B}  (2022/09/11 $\sim$ 2022/09/20) and \textbf{Sep. C}  (2022/09/21 $\sim$ 2022/09/30). The detailed statistics are presented in Table~\ref{tab:dataset}, in which we also roughly divide the queries into head and tail with exposure.

Based on the real-world dataset, we construct a unified service search graph and an intention tree for facilitating the learning of {\model}. As mentioned in Sec.~\ref{sec:graph_encode}, we employ two individual GNNs for encoding head and tail queries/services, respectively, driving us to make the head/tail queries split and correspondingly organize them as head and tail graphs in advance for performing adaptive encoding more conveniently.
Moreover, we extract about 11 semantic-related attributes for the service search graph, which is shared for all baselines. The detailed descriptions of the service search graph and intention tree are shown in Table~\ref{tab:dataset-graph}.  

\subsubsection{Public Datasets}
We also conduct an experiment on the well-known Amazon product dataset, which 
has become a benchmark dataset for evaluating product search methods in many recent studies~\cite{ai2019zero,ai2017learning,ai2019explainable}. 
Here, we choose 3 sub-datasets from different domains in our experiments, namely \textbf{Software}, \textbf{Video game} and \textbf{Music}. Specifically, the Software dataset includes 1,826 users and 802 items with 12,805 interactions, the Video game dataset includes 55,223 users and 17,408 items with 497,576 interactions, and the Music dataset includes 27,530 users and 10,620 items with 231,392 interactions. 
We follow a similar way in Sec.~\ref{sec-industrial} for the split of head/tail queries and the construction of a service search graph and an intention tree, and present the detailed descriptions of the public datasets in Table~\ref{tab:dataset} and Table~\ref{tab:dataset-graph}.

\input{tab/dataset}

\input{tab/dataset_graph}

\subsection{Experimental Setup}
Here, we present experimental settings, including compared baselines, evaluation metrics and implementation details.

\subsubsection{Baselines}
Considering the industrial settings, the selected baselines have potential for scaling up to a huge volume of datasets and meeting the urgent requirements of low delay in online serving. Therefore, we compare {\model} with three types of state-of-the-art methods, covering general deep neural network models (\ie \textbf{Wide\&Deep}~\cite{cheng2016wide}), GNN based models (\ie \textbf{LightGCN}~\cite{he2020lightgcn} and \textbf{KGAT}~\cite{wang2019kgat}) and GNN based models with self-supervised learning (\ie \textbf{SGL}~\cite{wu2021self} and \textbf{SimSGL}~\cite{yu2022graph}). 
It is worthwhile to note that rich attributes on nodes and edges are involved in our service search graph, and thus we extend LightGCN, SGL and SimSGL to adapt to it for more promising and fair results. 
\subsubsection{Evaluation metrics}
We adopt three widely-used metrics in the field of recommendation and search,
namely \textbf{AUC}, \textbf{GAUC} and \textbf{NDCG@K}~\cite{he2017neural,shi2019deep}.

\input{tab/eff_exp_table}

\input{tab/eff_exp_table_2}

\subsubsection{Implementation details}
For scaling up to large-scale datasets adopted in the paper, {\model} and all compared methods are implemented on our parameter server based distributed learning systems.
For fair comparison, we set batch size = $1024$, embedding size = $64$ and select ADAM optimizer with learning rate = $1e-4$ for all methods. For all baselines, we refer to  the optimal configuration and architecture reported in the original literature and carefully tune all parameters in the validation set. For reproducibility, we list the parameter values used in our model as follows: For all datasets, we set $H = 5, L = 2, \alpha = 0.1, \beta = 0.01, \tau = 0.1$. An in-depth analysis of the key hyper-parameters will be presented in Sec.~\ref{sec:para}.

\subsection{Overall Performance Compared with Baselines (\textbf{RQ1})}
Comprehensive performance comparison \wrt AUC of {\model} and the baselines on different query slices are reported in Table~\ref{tab:baseline}. And we also report the experimental results \wrt GAUC and NDCG@10 on tail queries in Table~\ref{tab:baseline_2}. Overall, we find that {\model} generally brings the improvement \wrt AUC for both tail and head queries, as well as the overall performance, which is a highly encouraging result since improving both tail and head queries at the same time is non-trivial. In detail, we have the following key observations:
\begin{itemize}[leftmargin=*]
    \item Compared with the general deep neural model - Wide\&Deep, a popular baseline in industrial applications, the {\model} generally gains great improvements for both tail and head queries. The results demonstrate the crucial importance of carefully considering the long-tail issue in the service search scenarios, as well as the usefulness of graph structure for enriching tail queries. We also discover the larger performance margin between {\model} and  Wide\&Deep \wrt GAUC and NDCG@10 on the tail queries, further verifying the superior capacity of {\model} for powering  representations of tail queries.
    \item Compared with GNN based models (\ie LightGCN and KGAT), {\model} significantly and consistently outperforms them by a considerable margin across all the metrics of all the datasets.
    Since head queries/services are commonly paid disproportionate attention in the training procedure of {LightGCN and KGAT}, the patterns of tail queries/services are not well understood due to their minority. On the contrary, {\model} is equipped with a well-designed multi-granularity
    contrastive learning, greatly benefiting the learning of tails with auxiliary supervision.
    Moreover, the improvement of KGAT and lightGCN \wrt Wide\&Deep indicates the necessity of incorporation of structural information, while the improvement of KGAT \wrt LightGCN reveals that the neighbors for the target queries/services in the graph should be carefully weighted (or selected). 
    \item Compared with the GNN based models with self-supervised learning (\ie SGL and SimSGL), {\model} always achieve better performance in all cases. Since both baselines and {\model} share the same GNN encoder, we argue that such an encouraging performance gain is attributed to the following two promising designs: 
    i) We separately equip the head and tail queries/services with individual GNN encoders, guiding {\model } to pay more attention to the tail queries/services. 
    ii) We creatively devise the multi-granularity contrastive learning,  enhancing the expressiveness of tails through knowledge transfer, structure enhancement and intention generalization.
    It is noteworthy that both SGL and SimSGL do not work well compared with LightGCN and KGAT. An intuitive explanation is that noises are widely exited in the structure and attribute of the industrial graphs, easily causing the failure of stochastic augmentation adopted in SGL and SimSGL. It also inspires us to explicitly consider meaningful relations (\eg structural relations among queries and hierarchical intention tree) for contrastive learning in {\model}. 
    
\end{itemize}

\begin{figure}[htp]  
	\begin{minipage}{0.24\textwidth}  
		\centerline{\includegraphics[width=1.0\textwidth]{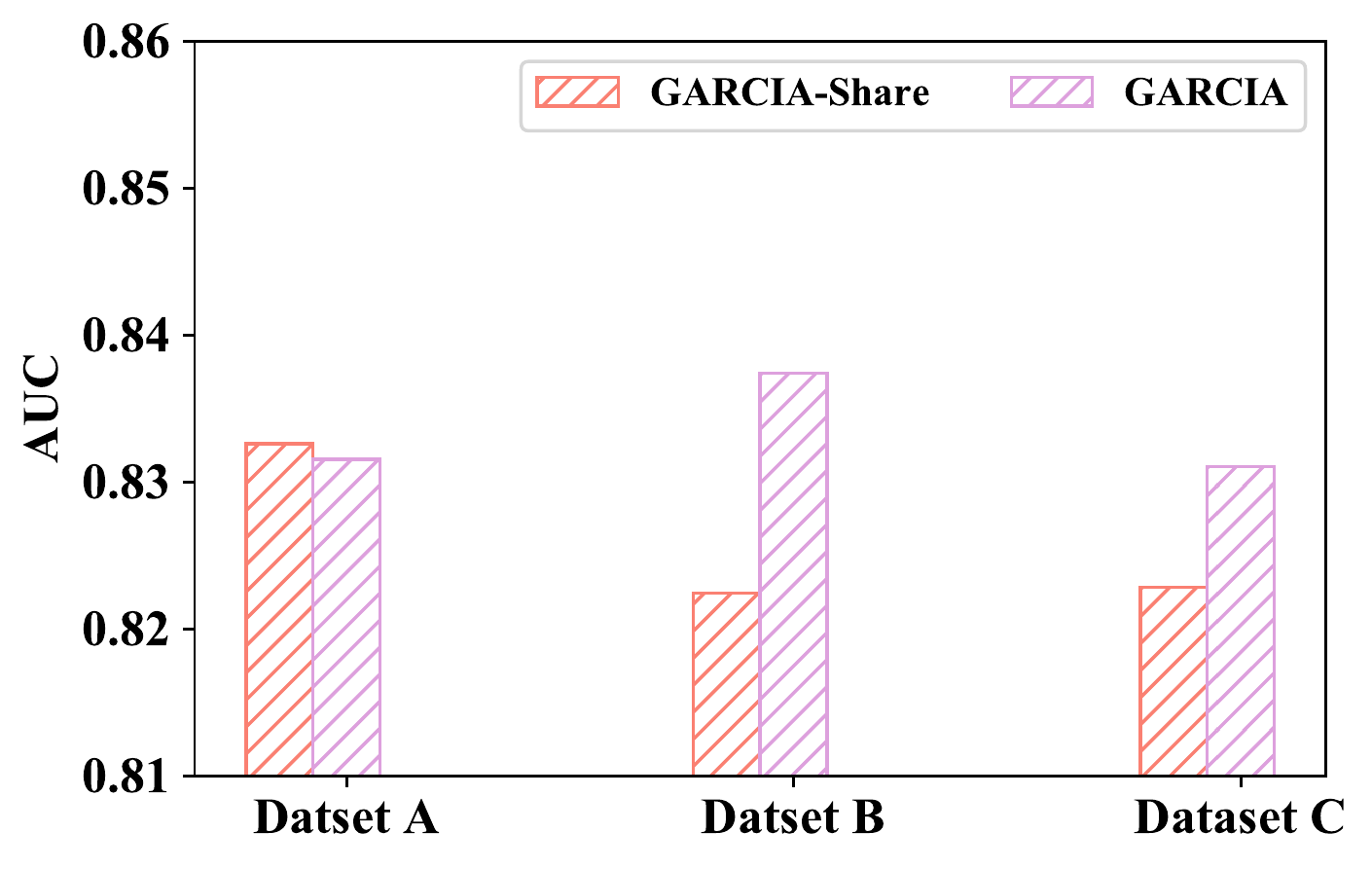}}
		\centerline{(a) Tail}
	\end{minipage}
	\hfill
	\begin{minipage}{0.24\textwidth}
		\centerline{\includegraphics[width=1.0\textwidth]{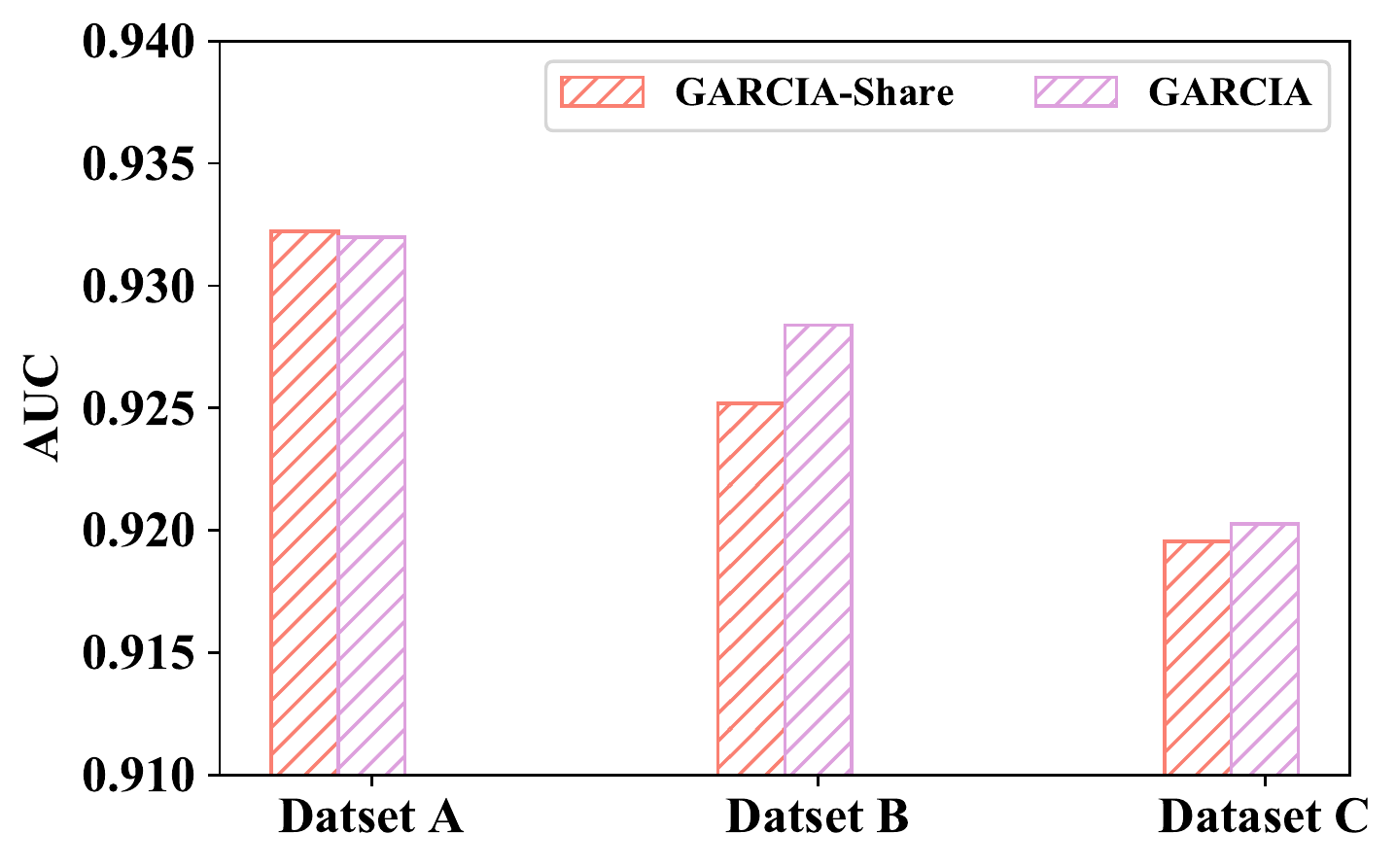}}
		\centerline{(b) Overall}
	\end{minipage}
	\caption{The ablation study on the adaptive encoding.}
	\label{fig:share_encoding}
\end{figure}

\begin{figure*}[htp]  
	\begin{minipage}{0.3\textwidth}  
		\centerline{\includegraphics[width=1.1\textwidth]{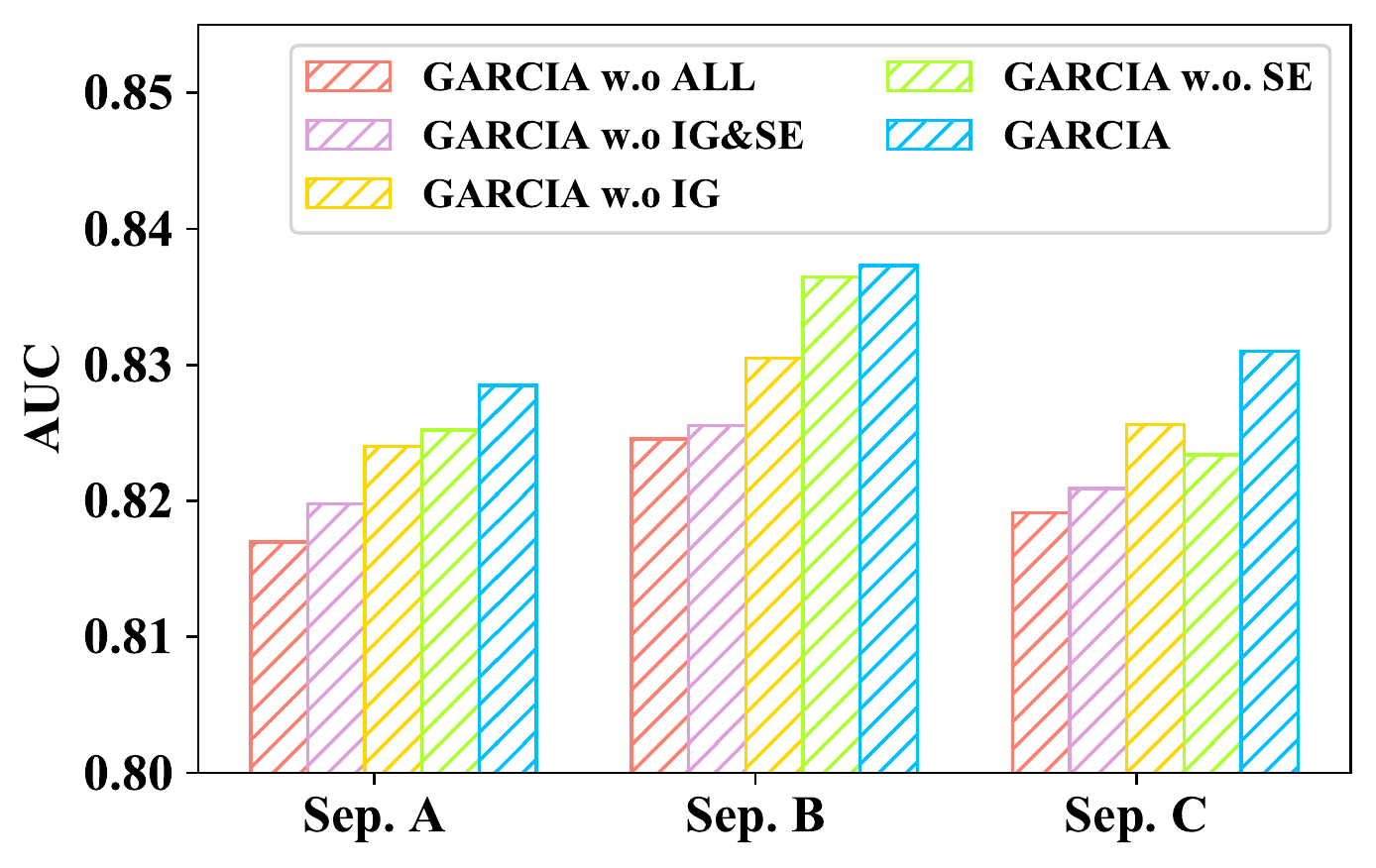}}
		\centerline{(a) Tail}
	\end{minipage}
	\hfill
	\begin{minipage}{0.3\textwidth}
		\centerline{\includegraphics[width=1.1\textwidth]{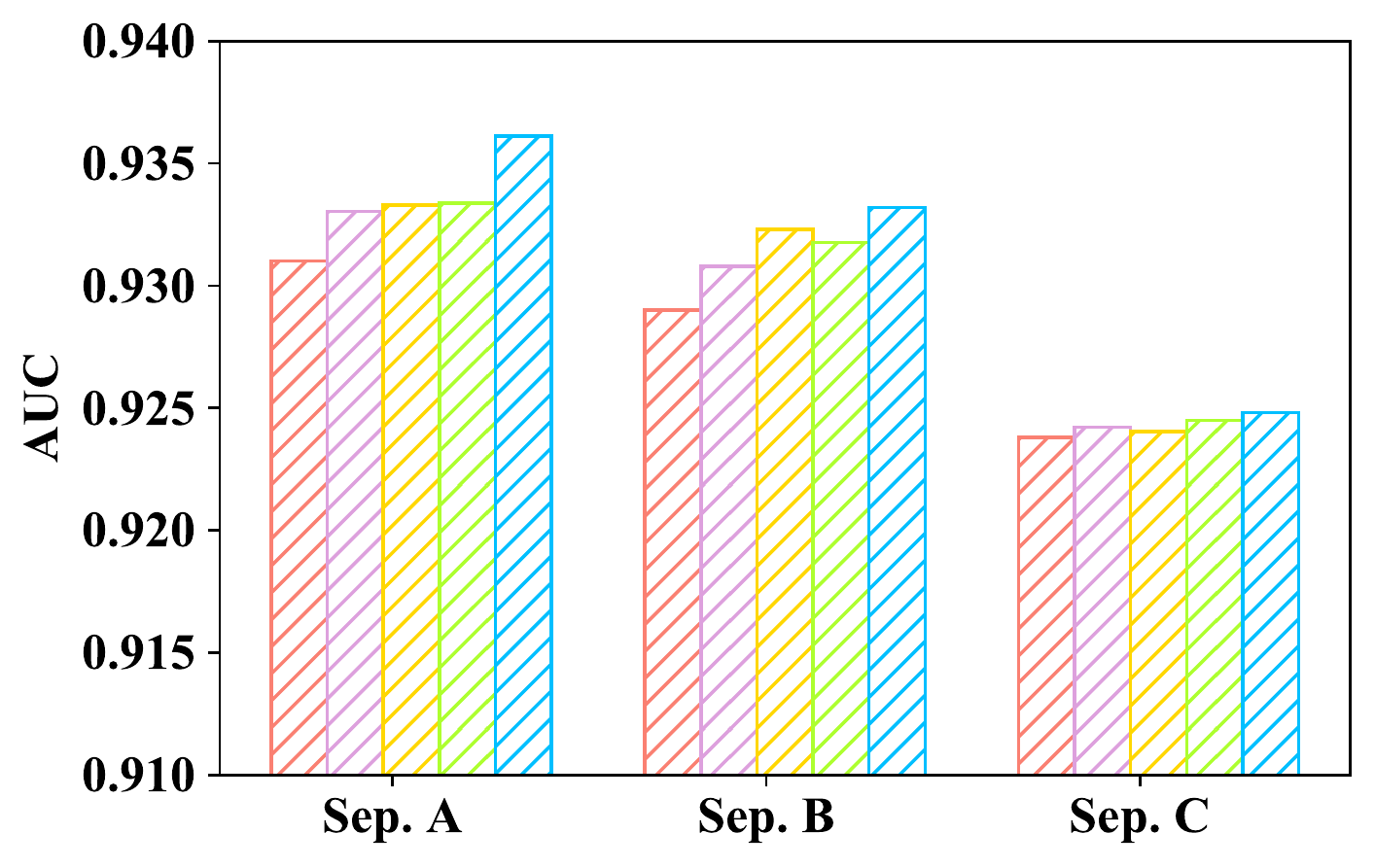}}
		\centerline{(b) Head}
	\end{minipage}
	\hfill
	\begin{minipage}{0.3\textwidth}
		\centerline{\includegraphics[width=1.1\textwidth]{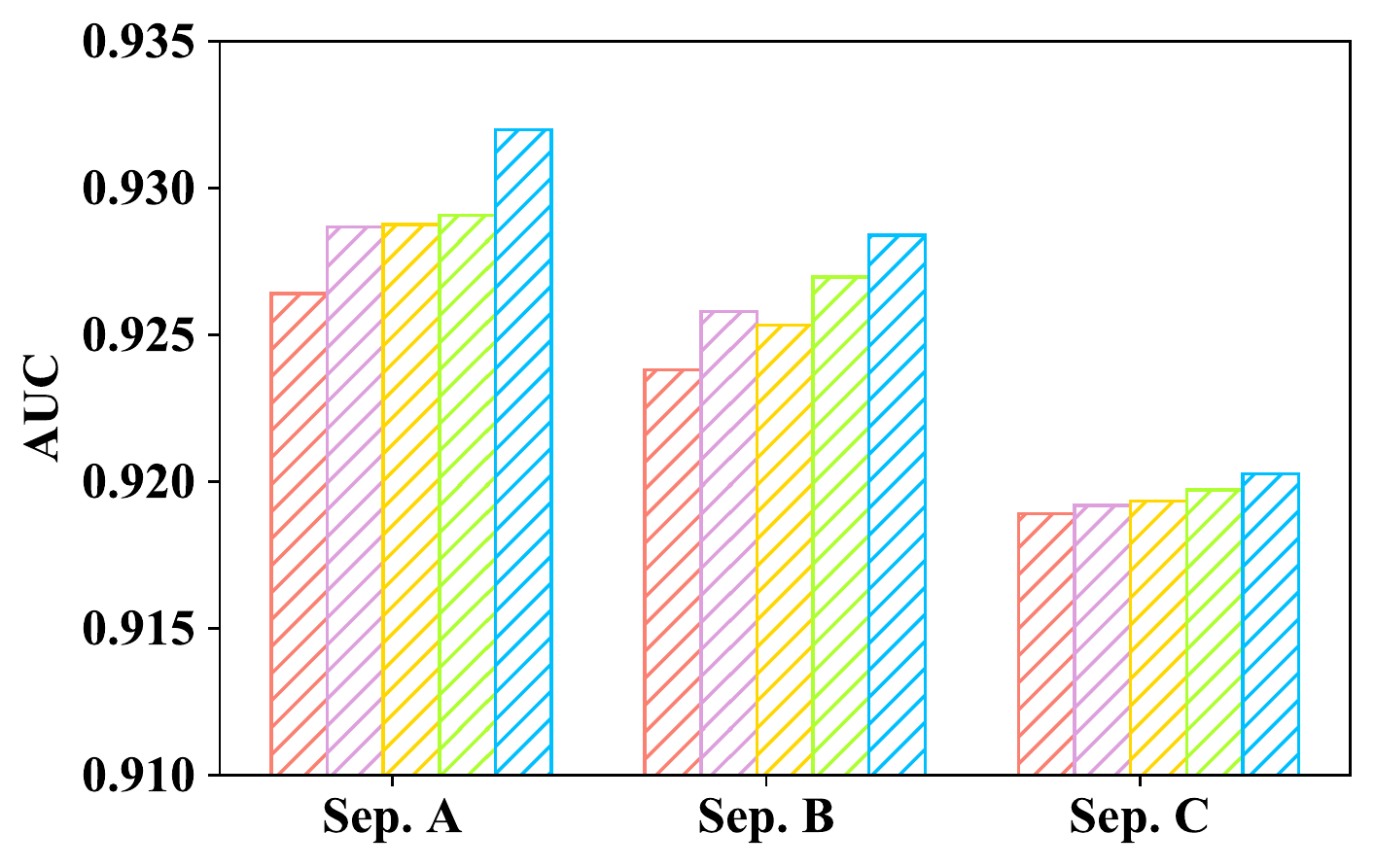}}
		\centerline{(c) Overall}
	\end{minipage}
	\caption{The ablation study on multi–granularity contrastive learning.}
	\label{fig:aba}
\end{figure*}

\subsection{Ablation Study (\textbf{RQ2})}
In this section, we conduct a series of ablation studies on {\model} to investigate the impact and rationality of the adaptive encoding for tails and heads, and the multi-granularity contrastive learning component.

\subsubsection{Adaptive encoding for tails and heads}
First, we investigate the adaptive encoding for tails and heads, which adopt two individual GNNs for representation learning on the tail and head queries/services, respectively. Hence, we prepare a variant of {\model}, named \textbf{{\model}-Share}, in which a unified GNN is shared for tails and heads.

As shown in Fig.~\ref{fig:share_encoding}, we observe {\model} achieves comparable performance with {\model}-Share in Sep. A, while {\model} outperform {\model}-Share by a considerable margin in both Sep. B and C. The results demonstrate the  superiority of encoding tails and heads in an adaptive manner, which helps {\model} shift attention towards structural patterns of tail queries/services, and thus benefit the alleviation of the long-tail issue in the service search scenario.

\subsubsection{Multi-granularity contrastive learning}
To comprehensively understand  contributions of multi-granularity contrastive learning towards {\model} for tackling the long-tail issue in the service search, we prepare the following four variants of {\model}:
i) \textbf{{\model} w.o. SE}: {\model} without the SECL (\ie Eq.~\ref{eq:secl}).
ii) \textbf{{\model} w.o. IG}: {\model} without the IGCL (\ie Eq.~\ref{eq:igcl}).
iii) \textbf{{\model} w.o. ALL}: {\model} without multi-granularity contrastive learning, consisting of 
iiii)  \textbf{{\model} w.o. ALL}: {\model} without multi-granularity contrastive learning, consisting of 
    the KTCL (\ie Eq.~\ref{eq:ktcl}), the SGCL (\ie Eq.~\ref{eq:secl}) and IGCL (\ie Eq.~\ref{eq:igcl})

We report the experimental results in Fig.~\ref{fig:aba}, from which we have the following two key findings 
i) The performance will drop a lot if the multi-granularity contrastive learning is removed (\ie \textbf{{\model} w.o. ALL}), indicating the effectiveness of  multi-granularity contrastive learning. 
ii) Ignoring any contrastive supervision is not ideal (\ie  \textbf{{\model} w.o. SE}, \textbf{{\model} w.o. IG}, \textbf{{\model} w.o. IG\&SE}), disclosing that both of knowledge from heads, local structure and intention tree have their own contributions to the learning of tails.

\begin{figure}[htp]  
	\begin{minipage}{0.24\textwidth}  
		\centerline{\includegraphics[width=1.1\textwidth]{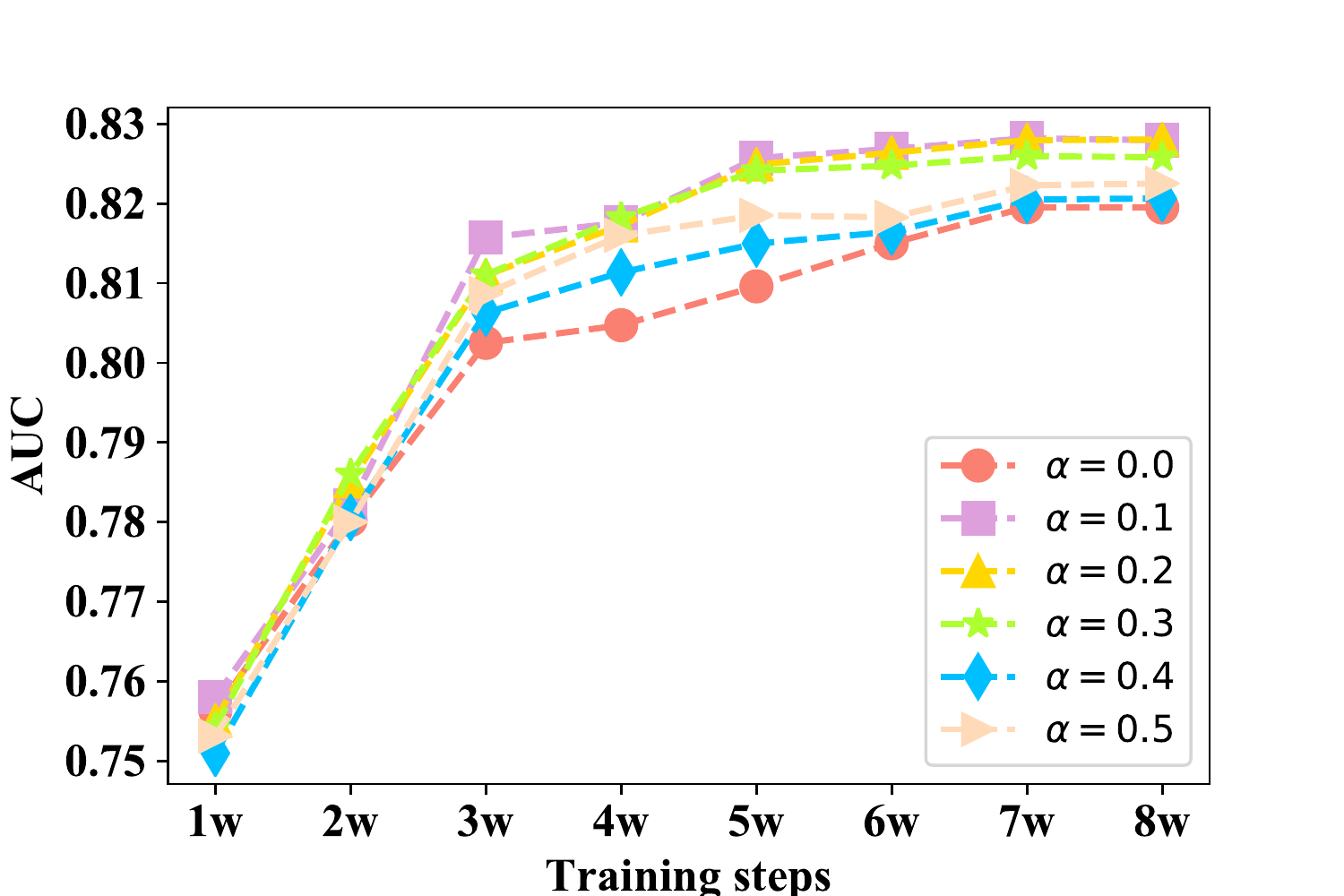}}
		\centerline{(a) Tail}
	\end{minipage}
	\hfill
	\begin{minipage}{0.24\textwidth}
		\centerline{\includegraphics[width=1.1\textwidth]{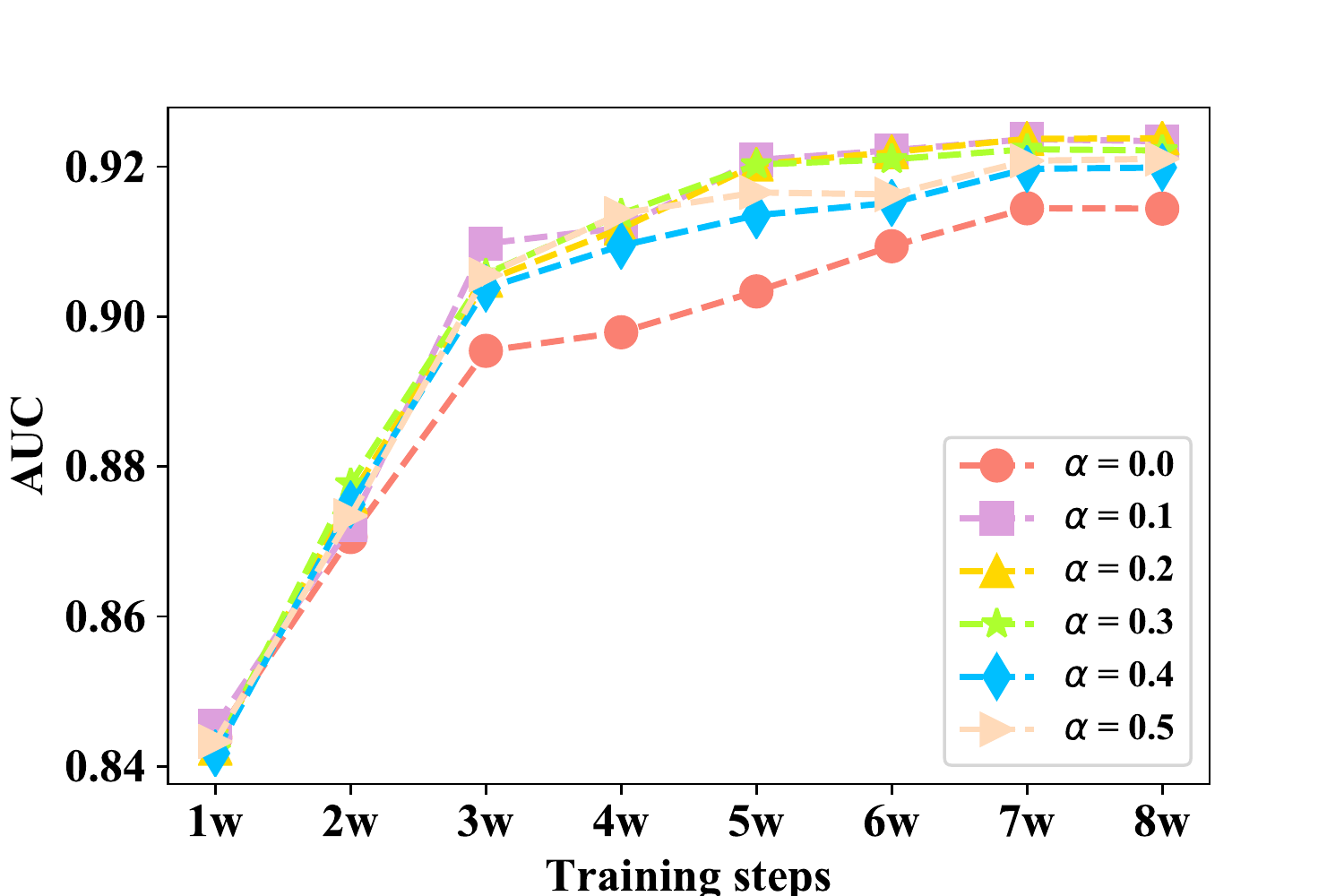}}
		\centerline{(b) Overall}
	\end{minipage}
	\caption{The performance effect of the varying of the parameter $\alpha$, which shows that the optimal performance is achieved when $\alpha = 0.1 \sim 0.3$. }
	\label{fig:neighbor_cl}
\end{figure}

\begin{figure}[htp]  
	\begin{minipage}{0.24\textwidth}  
		\centerline{\includegraphics[width=1.1\textwidth]{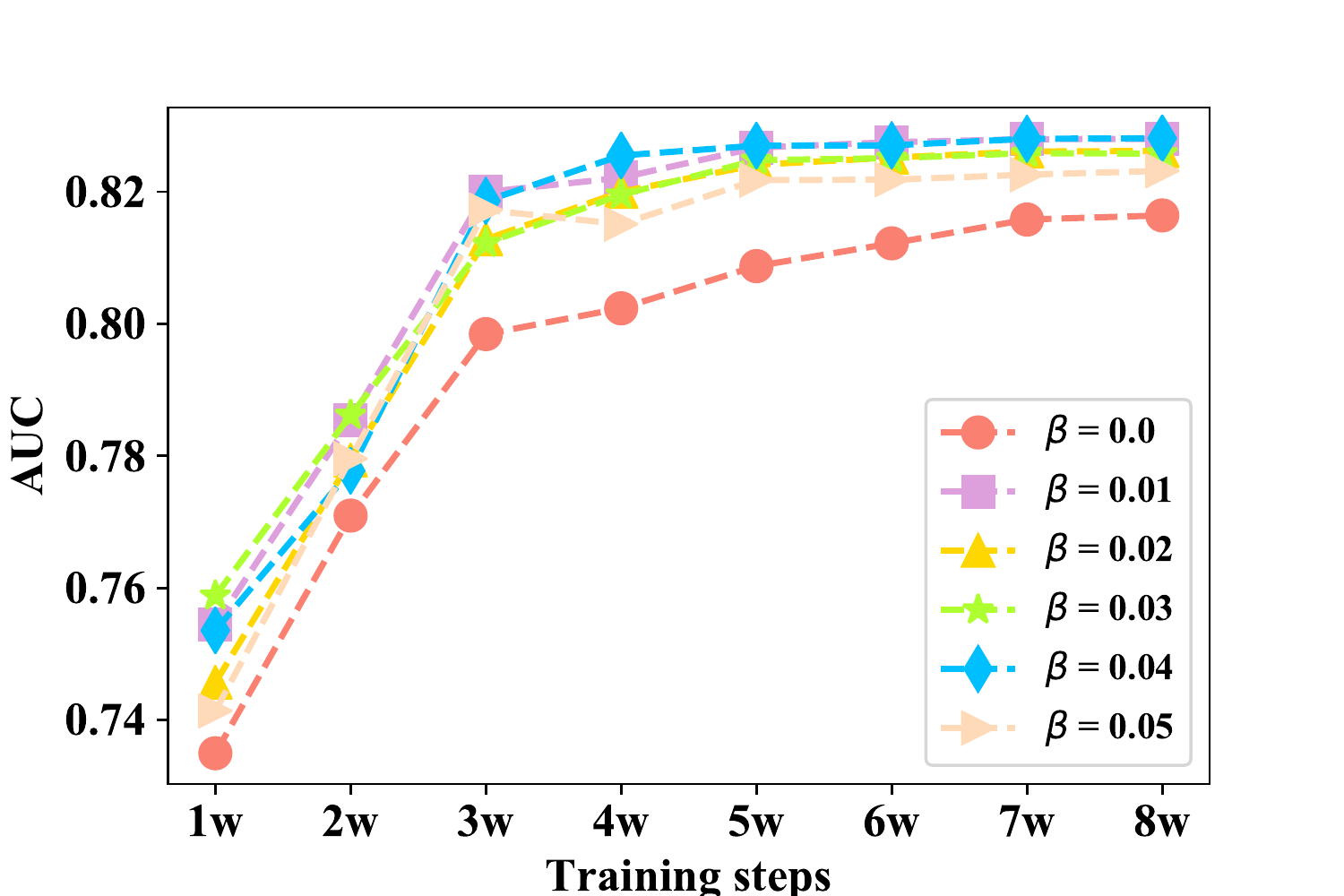}}
		\centerline{(a) Tail}
	\end{minipage}
	\hfill
	\begin{minipage}{0.24\textwidth}
		\centerline{\includegraphics[width=1.1\textwidth]{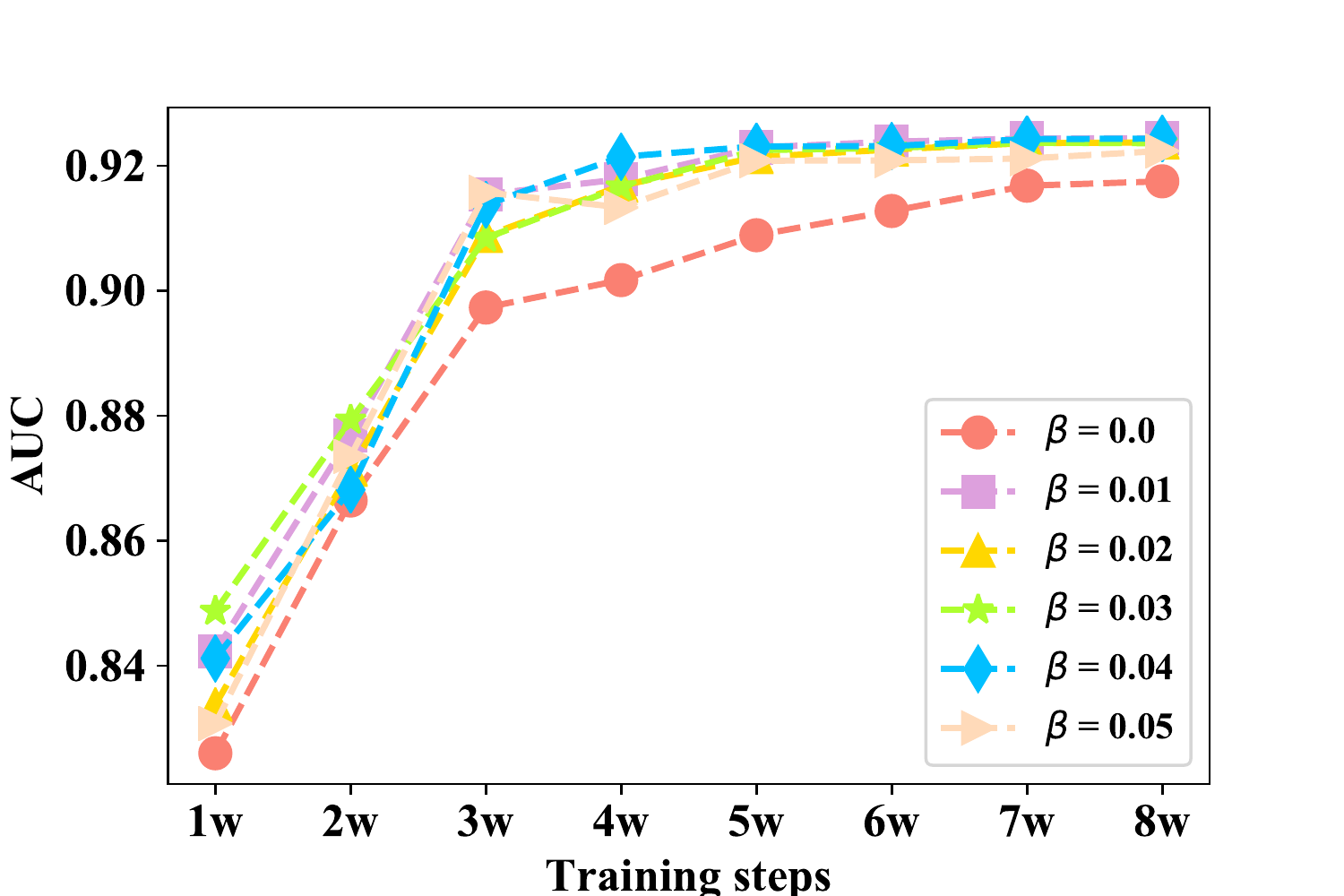}}
		\centerline{(b) Overall}
	\end{minipage}
	\caption{The performance effect of the varying of the parameter $\beta$, which shows that our approach achieves the best performance when $\beta = 0.01$ or $0.04$. }
	\label{fig:intent_cl}
\end{figure}

\begin{figure}[htp]  
	\begin{minipage}{0.24\textwidth}  
		\centerline{\includegraphics[width=1.1\textwidth]{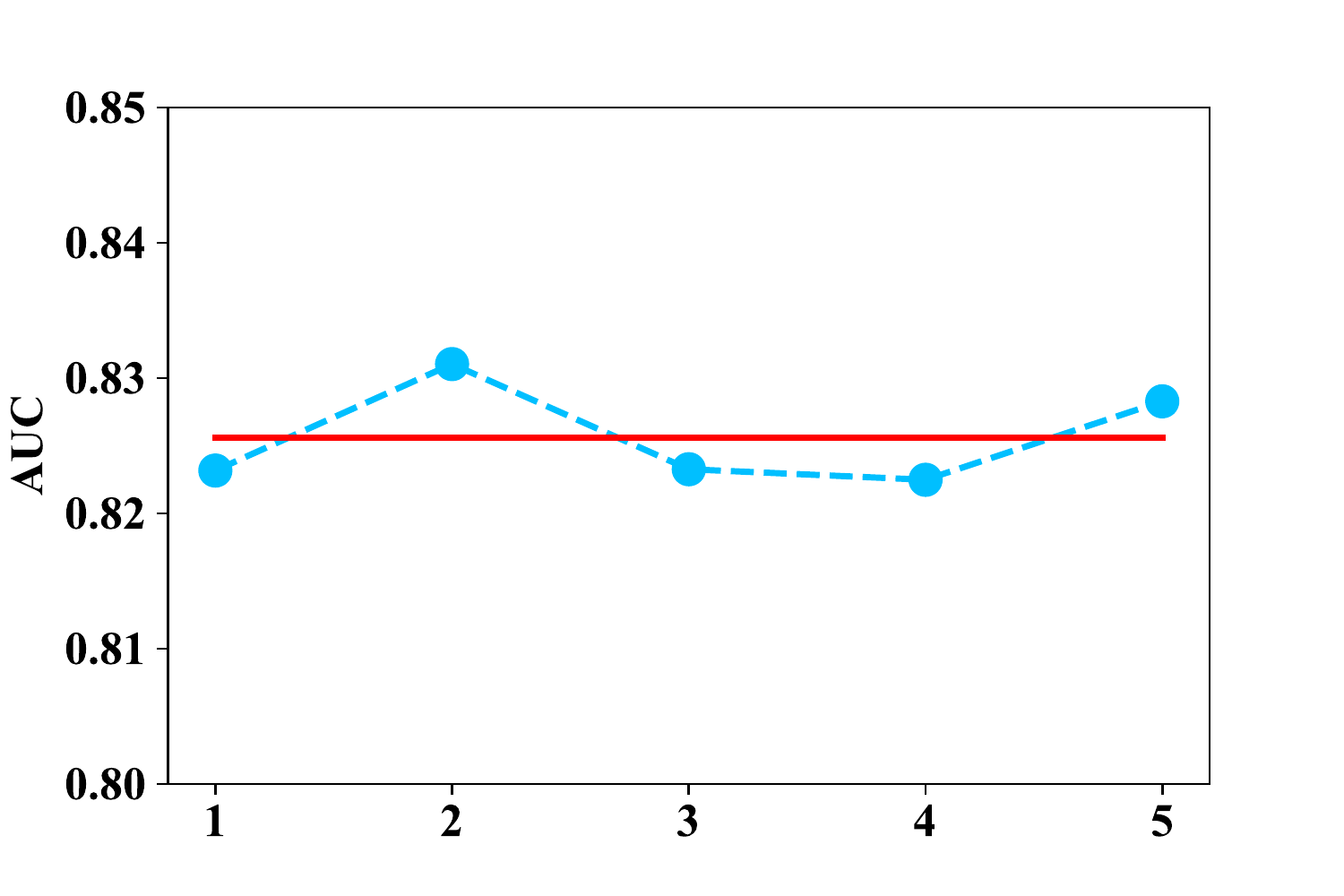}}
		\centerline{(a) Tail}
	\end{minipage}
	\hfill
	\begin{minipage}{0.24\textwidth}
		\centerline{\includegraphics[width=1.1\textwidth]{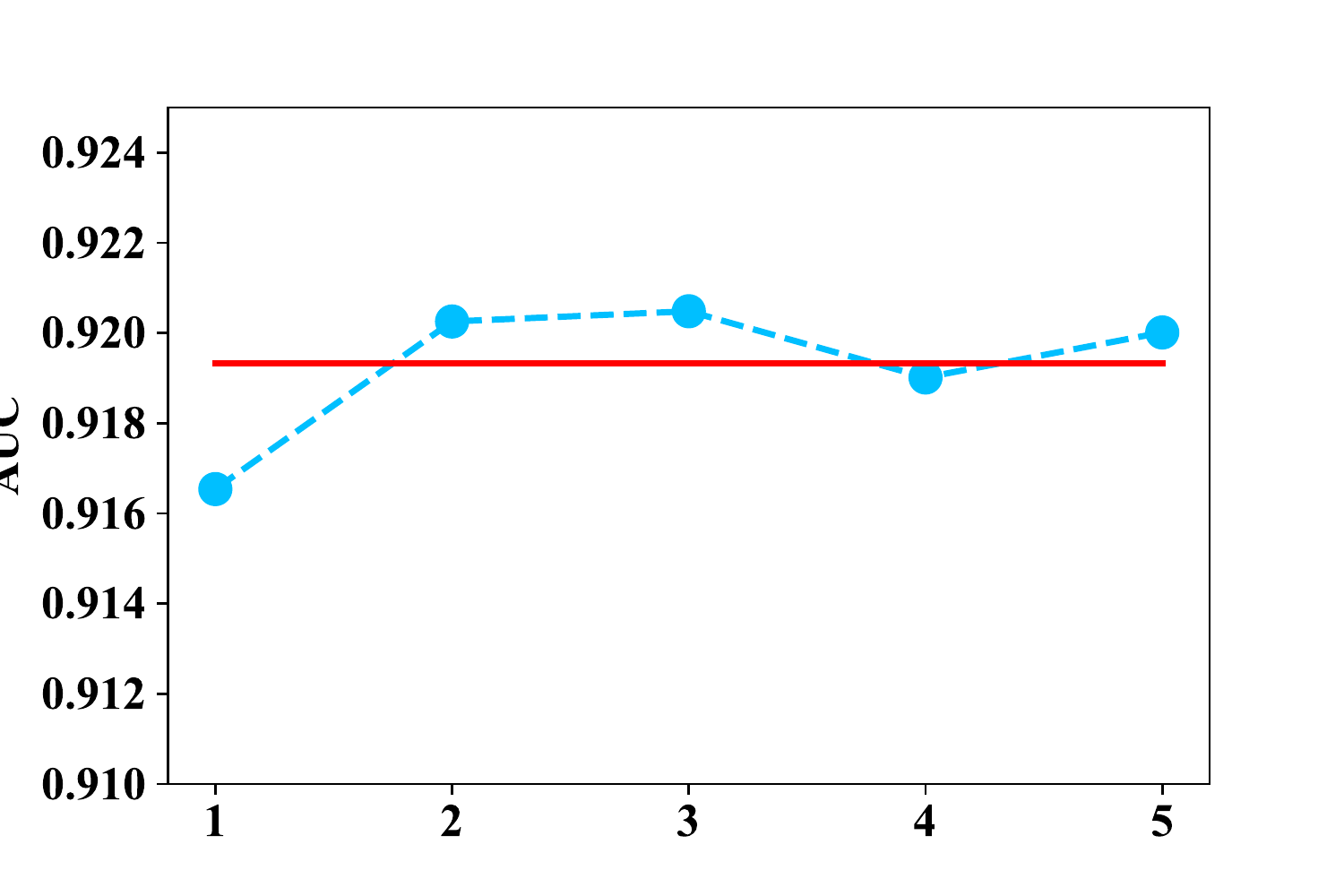}}
		\centerline{(b) Overall}
	\end{minipage}
	\caption{Impact of the number of levels of intention tree H. We involve the GARCIA without consideration of intention as the reference baseline (\ie the red solid line).}
	\label{fig:intent_aba}
\end{figure}

\begin{figure}[htp]  
	\begin{minipage}{0.24\textwidth}  
		\centerline{\includegraphics[width=1.1\textwidth]{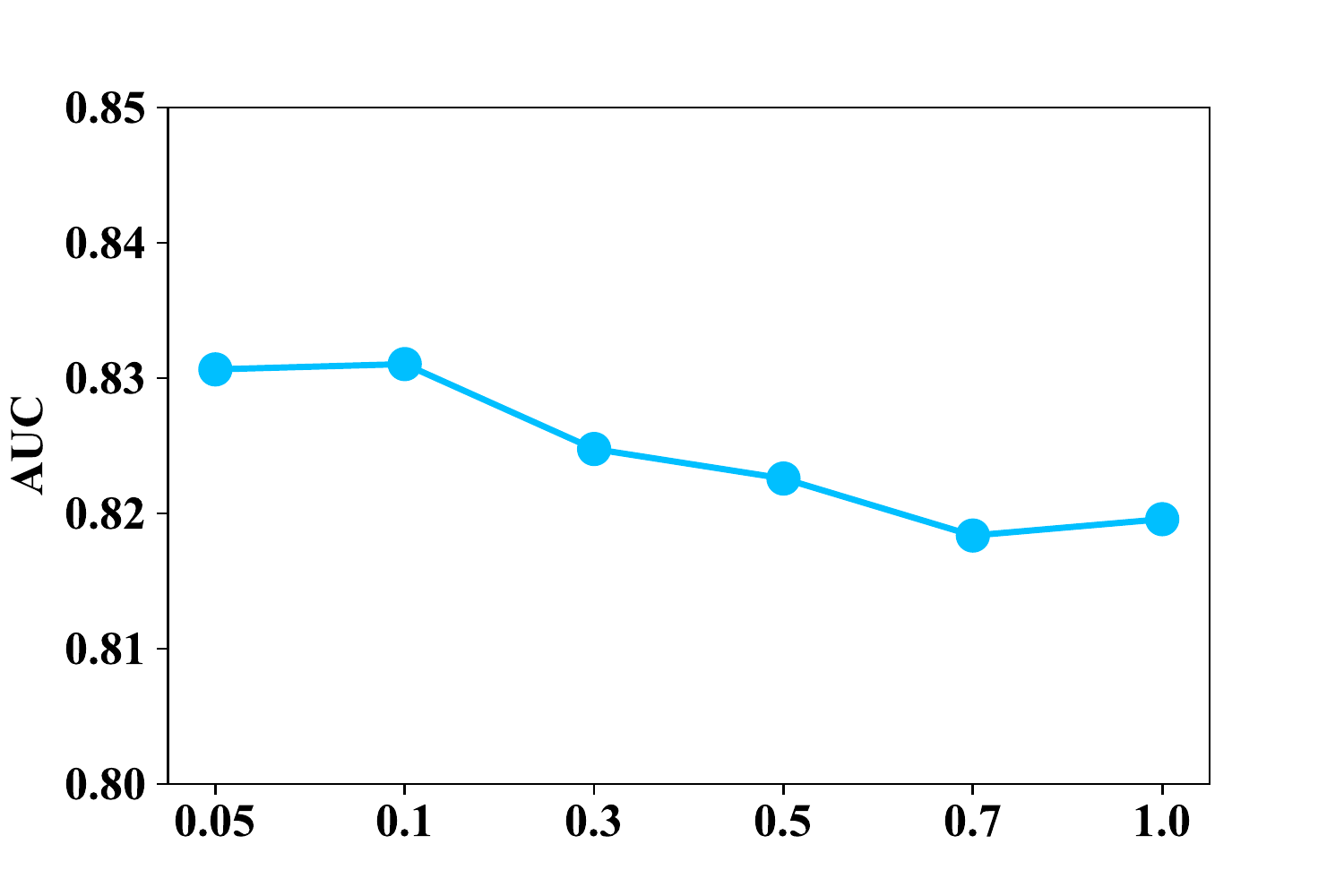}}
		\centerline{(a) Tail}
	\end{minipage}
	\hfill
	\begin{minipage}{0.24\textwidth}
		\centerline{\includegraphics[width=1.1\textwidth]{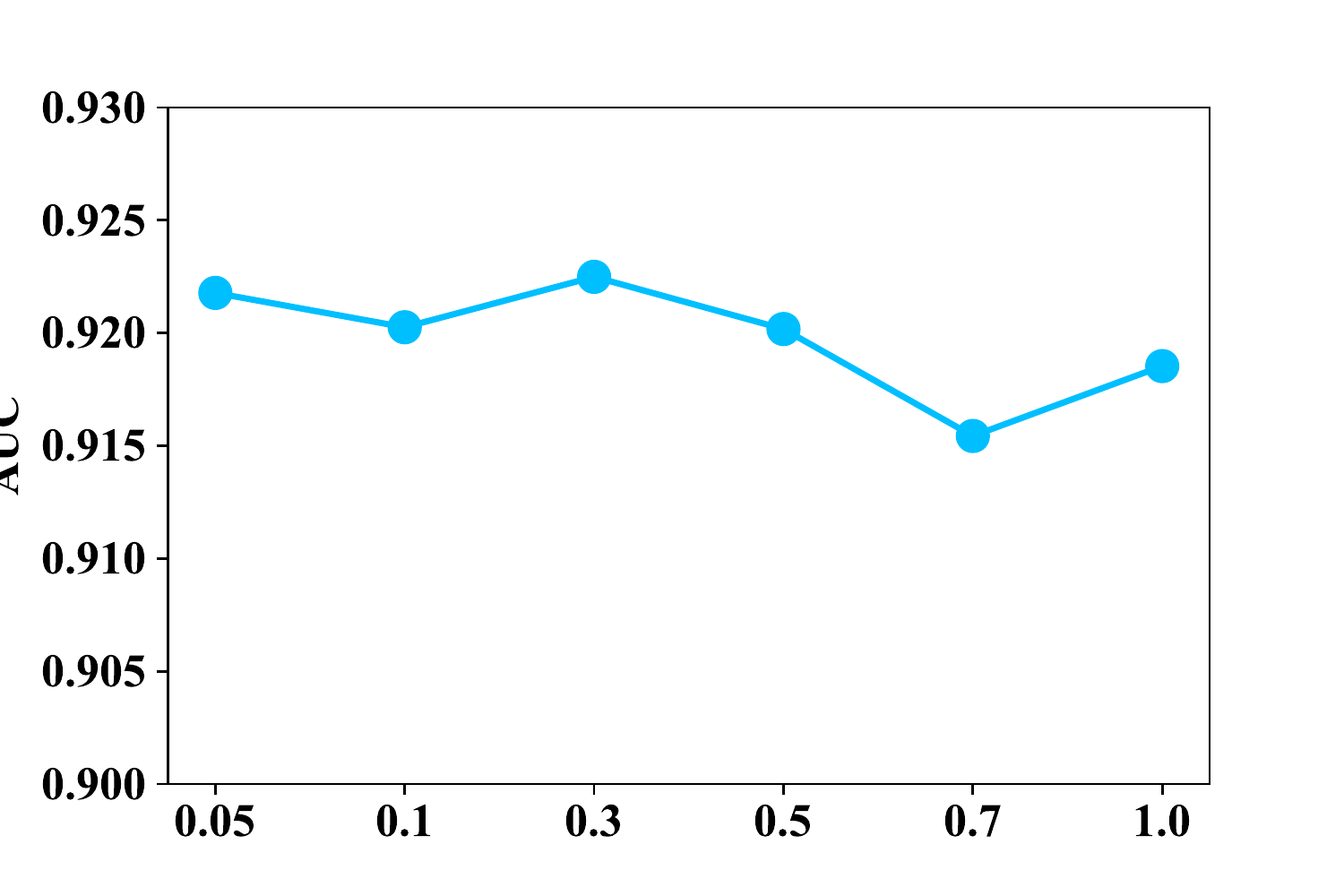}}
		\centerline{(b) Overall}
	\end{minipage}
	\caption{Parameter tuning of $\tau$ in contrastive learning, which reveals that $\tau$ cannot be set too large.}
	\label{fig:tau}
\end{figure}

\subsection{Hyper-parameter Sensitivity (\textbf{RQ3}) \label{sec:para}}
As mentioned above, {\model} involves several important hyper-parameters, \ie balance factors $\alpha$ and $\beta$ in $\mathcal{L}_\mathcal{P}$ (\ie Eq. 11), the levels of intention tree $H$ and the temperature factor $\tau$ in each contrastive loss.
Here, we investigate the impact of these parameters in the following categories, using the ranking quality on the tail and overall queries in the Sep. A dataset. Similar patterns have been observed in the performance of remaining datasets (\ie Sep. B and C).

\subsubsection{Impact of the balance factors $\alpha$ and $\beta$}
In Eq. 11, $\alpha$ and $\beta$ control the weight of $\mathcal{L}_{SECL}$ and $\mathcal{L}_{IGCL}$, and we examine the performance effect of two parameters  with the grid search strategy.
Concretely, we vary $\alpha$ in the set of \{ 0.0, 0.1, 0.2, 0.3, 0.4, 0.5\} and $\beta$ in \{ 0.0, 0.01, 0.02, 0.03, 0.04, 0.05\}, respectively~\footnote{We experimentally find that {\model} always yields relatively poor performance when $\alpha > 0.5$ or  $\beta > 0.05$}. 
As shown in Fig.~\ref{fig:neighbor_cl} and Fig.~\ref{fig:intent_cl}, we find: 
i) {\model} achieve worst performance when $\alpha = 0$ or $\beta = 0$, demonstrating the indispensability of SECL and IGCL.
ii) A large $\alpha$ or $\beta$ will cause the sharp performance degradation, and the optimal performance is obtained when $\alpha = 0.1 \sim 0.3$ and $\beta = 0.01$ or $0.04$.

\subsubsection{Impact of the number of levels of intention tree $H$}
In our scenarios, the maximum level of an intention sub-tree is 5, and thus, we conduct a sensitivity analysis incorporating different numbers of levels of the intention tree (\ie 1 $\rightarrow$ 5).
In Fig.~\ref{fig:intent_aba}, we can observe that generally the performance {\model} improves with more levels of intention tree incorporated. Whereas it does not always yield improvement, and a slight performance fluctuation appears when $H = 3$ or $H = 4$, which is possibly attributed to the underlying noise in the intention tree. On the other hand, we can effectively control the model complexity by carefully tuning the parameter $H$.

\subsubsection{Impact of temperature factor $\tau$}
Finally, we analyze the impact of temperature factor $\tau$, which plays important role in contrastive learning. Similarly, we vary it in the set of \{0.05, 0.1, 0.3, 0.5, 0.7, 1.0 \} and report the performance change in the Fig.~\ref{fig:tau}. From the figure, we find that {\model} achieves optimal performance when $\tau = 0.1$ and is generally stable around that value, and too large values would harm the model.

\subsection{Performance in the Online Environment (\textbf{RQ4})}

\subsubsection{Online deployment}
\begin{figure}
  \centering
  \includegraphics[width=0.5\textwidth]{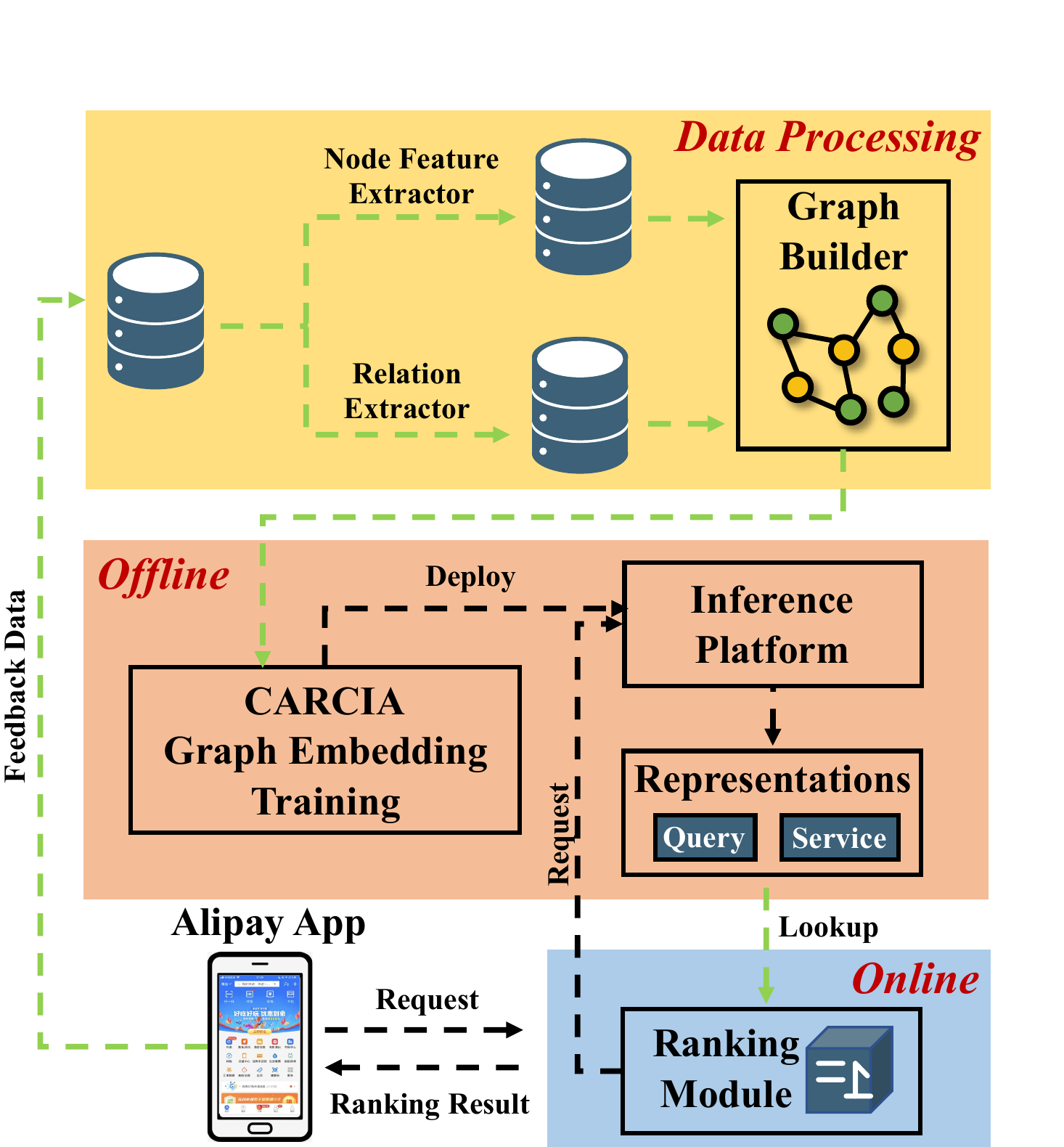}
  \caption{The deployment details of {\model} in the service search scenario of Alipay.}
  \label{fig:search application System}
\end{figure}

To further verify the effectiveness of the proposed {\model}, we carefully deploy it in the service search scenario of Alipay, serving hundreds of thousands of users every day. As shown in Fig.~\ref{fig:search application System}, considering the low delay for online inference, {\model} is deployed in a hybrid offline-online manner, which could be roughly summarized as such a pipeline: data processing $\rightarrow$ offline training $\rightarrow$ online serving. 

The top of Fig.~\ref{fig:search application System} describes the construction details of the service search graph, which is supported by the \emph{Node Feature Extractor} and \emph{Relation Extractor}. In this way, abundant features associated with queries/services and high-quality relation meeting the criterion of interaction and correlation (See Section.~\ref{sec:pre}) are effectively mined in an automatic manner.
Then the well-established service search graph, coupled with training samples is fed into {\model} for model learning. 
It is worthwhile to note that we replace the MLP operation in Eq.~\ref{eq:pred} with a simple inner product operation for {\model} in the online environment. The underlying reason is inner product operation could help {\model} perform efficient embedding retrieval to avoid amounts of delays for online real-time inference.
After we deploy the well-trained {\model} in the \emph{Inference Platform}, embedding inference for queries and services is daily executed for online serving.
When it comes to the online serving, once a new-coming user issue a request, efficient embedding retrieval and similarity calculation are successively employed in the \emph{Ranking Module}, and our system only keeps top $K$ services with the highest similarities to form the final ranking results.

\begin{figure}[htp]  
	\begin{minipage}{0.24\textwidth}  
		\centerline{\includegraphics[width=1.0\textwidth]{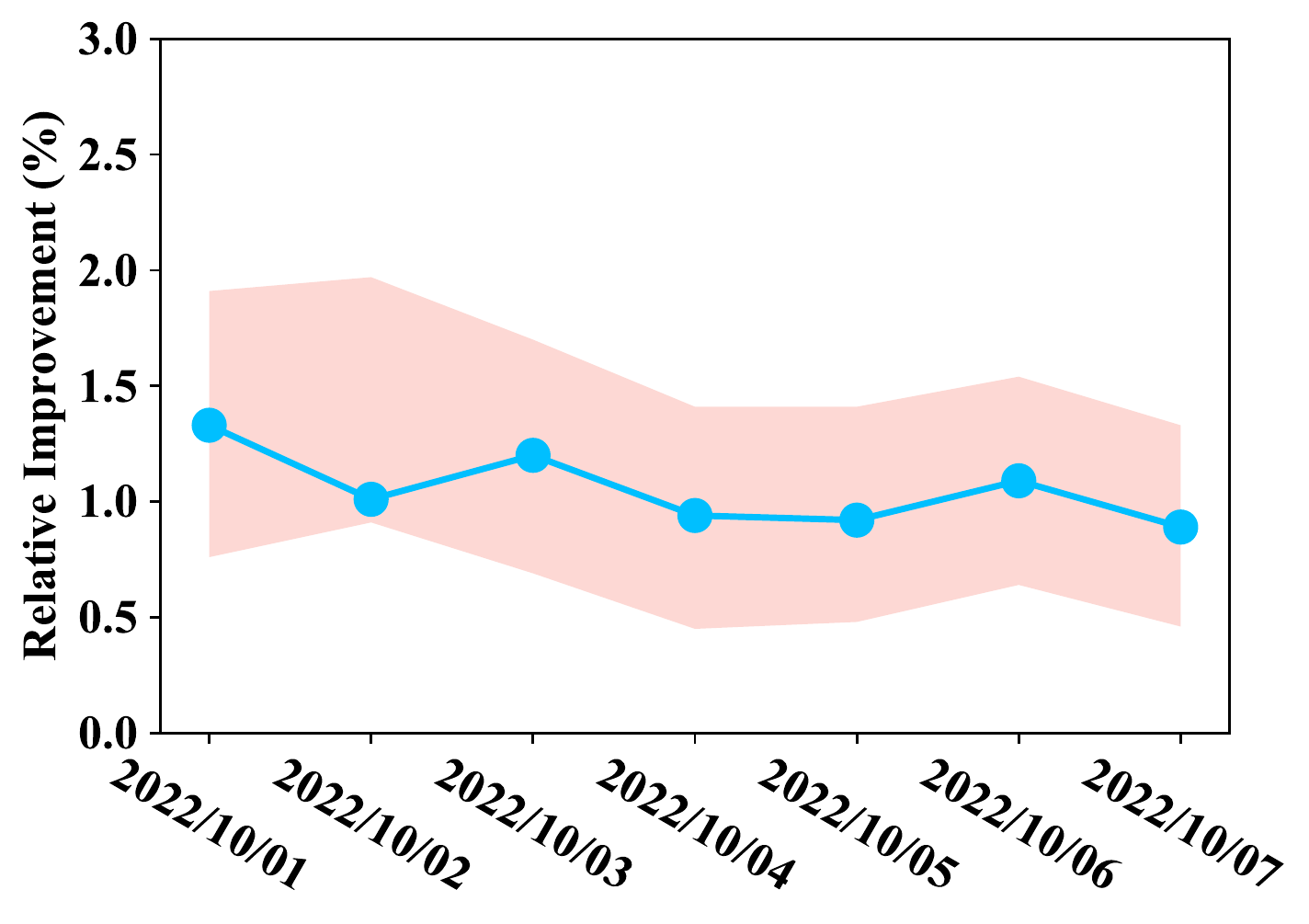}}
		\centerline{(a) CTR}
	\end{minipage}
	\begin{minipage}{0.24\textwidth}
		\centerline{\includegraphics[width=1.0\textwidth]{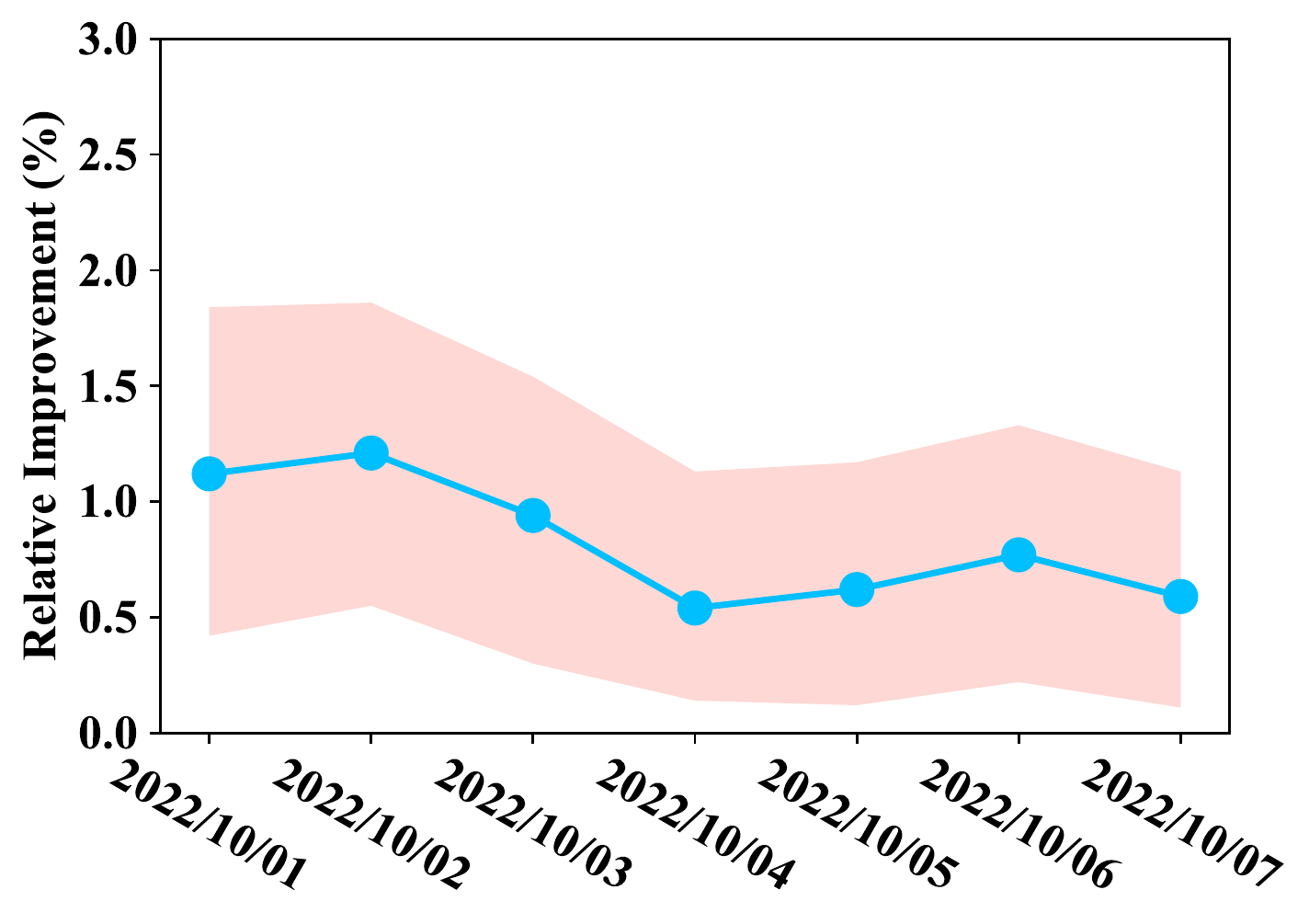}}
		\centerline{(b) Valid CTR}
	\end{minipage}
	\caption{Online performance on real-world service search scenarios in Alipay. $y$-axis denotes the improvement ratio over the deployed baseline.}
	\label{fig:online}
\end{figure}

\begin{figure*}[t]
  \includegraphics[width=1.0\textwidth]{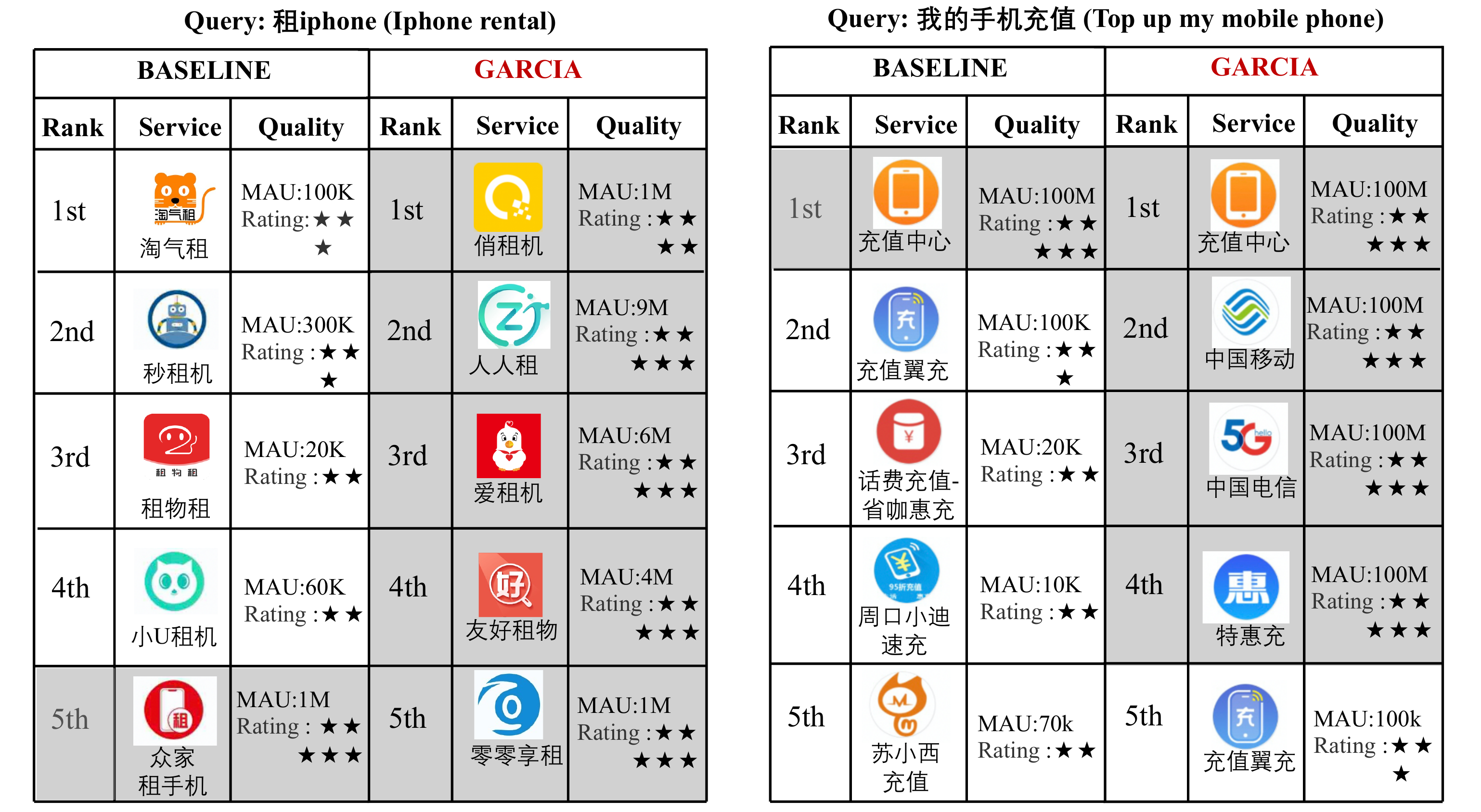}
  \caption{Examples of service retrieval with the specific query by the deployed baseline and the proposed {\model}. We mark the services with high quality with shadow. Best viewed in color.}
  \label{fig:real_case}
\end{figure*}

\subsubsection{Online performance}
To  evaluate the proposed {\model} in the online platform, we conduct online experiments in the service search scenario of Alipay. In particular, we conduct the bucket testing (\ie A/B testing) online to test the users’ response to our model against the baseline. One bucket is selected for the baseline (\ie the KGAT augmented Wide\&Deep model), 
and another is selected for our model. We utilize the metric \textbf{CTR} and \textbf{Valid CTR} for the online performance evaluation, where \textbf{CTR} only means the probability that a user clicks a service, while \textbf{Valid CTR} means the probability that a user clicks inside the service (similar to the concept of CVR, \ie ConVersion  Rate).
We perform the online evaluation from ``2022/10/01” to ``2022/10/07” and report the experimental results in Fig.~\ref{fig:online}.

From the experimental results, we observe that {\model} achieves consistent and significant performance improvement in all cases. Overall, {\model} gains the absolute improvement of {0.79\% and 0.60\%} \wrt CTR and Valid CTR with the significant statistics. This practice-oriented experiment further verifies that incorporating multi–granularity contrastive learning can better ease the long-tail issue, and thus  greatly improve the user’s search experience.


\subsubsection{Case study}
At last, we conduct case studies to show how {\model} improves the  rank quality of service for tail queries. Beforehand, we declare two dependable and intuitive metrics to measure the rank quality of services:
\begin{itemize}[leftmargin=*]
    \item \textbf{MAU} means the monthly active users of a certain service.
    \item \textbf{Authoritative Rating} is defined by our domain experts, which is a synthesis of several aspects for the target services, \eg whether it is a brand service, whether it is a worry-free service. 
\end{itemize}
In sum, besides the similarity between the query and service, we adopt the service popularity oriented metric (MAU) and the service quality oriented metric (Authoritative Rating) to measure the overall  
quality of service rank list provided by the deployed baseline and {\model}. 
Then, we are ready to take a closer look at the case studies, shown in Fig.~\ref{fig:real_case}:
\begin{itemize}[leftmargin=*]
    \item \textbf{Case I: long-tail query ``Iphone rental''}: The statistical analysis shows that more than 270,000 users search queries related to ``Mobile phone rental'' in Alipay every day, which brings more than 530,000 search PVs. On the comparison, the query ``Iphone rental'' is a representative long-tail query, whose rank list severely threatens the user experience. As shown in the left part of Fig.~\ref{fig:real_case}, not surprisingly, {\model} provides the rank list with higher MAU and authoritative ratings.

    \item \textbf{Case II: long-tail  query ``Top up my mobile phone''}: With the same intention,  the query ``Top up my mobile phone'' is a long tail query compared to the query ``Top up the mobile phone", where the latter is adopted by more than 130,000 users every day. As shown in the right part of Fig.~\ref{fig:real_case}, {\model} provides the more authoritative ranking results with  high-quality services, \eg  ``China Mobile'', ``China Telecom'', and ``Recharge with Special Discount''.
\end{itemize}
In sum, both of the cases intuitively demonstrate the superior capability of {\model} tailored for the long-tail issue in the service search scenarios.  

\input{sec-con}

\bibliographystyle{IEEEtran}
\bibliography{paper}

\end{document}

%% file: sec-rel.tex
\section{Related Works \label{sec:rel}}
In this section, we  review the most related works in graph-based search models, contrastive learning and long-tail issue.
\subsection{Graph-based Search Models} 
Recent years have witnessed the impressive success of graph neural networks(GNNs) in various applications, which aims at generalizing neural networks to graph structured data. Making full use of information propagation, graph-based product search seeks to organize interaction data (\ie ``user'' - ``query'' - ``item'') into a whole graph, and learn meaningful structure information through graph convolutions.
DHGAT~\cite{niu2020dual} proposes a dual heterogeneous graph attention network integrated with the two-tower architecture, using the user interaction data from both shop search and product search.
GEPS~\cite{zhang2019neural} employs graph embedding techniques to integrate graph-structured data into a unified neural ranking framework, which is the first study on how to use the click-graph features in neural models for retrieval.
SBG~\cite{fan2022modeling} proposes to explore local and global user behavior patterns on a user successive behavior graph, they also adopt a jumping graph convolution layer to alleviate the over-smoothing effect.
IHGNN~\cite{cheng2022ihgnn} constructs a hyper-graph from the ternary user-product-query interactions and applies graph neural networks to aggregate neighbor information on it.
DREM~\cite{ai2019explainable} constructs a directed unified knowledge graph based on both the multi-relational product data and the context of the search session, it jointly learns all embeddings through graph regularization. 
GraphSRRL~\cite{liu2020structural} explicitly utilizes specific user-query-product relationships which pay more attention to local relations. 

However, most of the graph-based approaches pay few attention to the long-tail queries, and neglect the hierarchical structure of intention among queries, which potentially benefits the generalization of query representations.

\subsection{Contrastive Learning} 
Contrastive learning (CL) has become a popular paradigm for self-supervised representation learning in the fields of natural language process and computer vision. With its promising capability for representation learning, several approaches come up with injecting CL into recommender systems to remedy the data sparsity problem~\cite{zou2022improving}. For instance, S$^3$-Rec~\cite{zhou2020s3} and DNN+SSL~\cite{yao2020self} commonly adopt a multi-task recommendation framework with the mutual information maximization principle. Recently, a series of studies marries the advantage of CL and GNNs for recommendation and attain impressive performance improvement. As a representative CL-based graph recommendation method, SGL~\cite{wu2021self} performs node and edge dropout to augment the original graph and adopts InfoNCE for CL upon the LightGCN framework, while SimGCL~\cite{yu2022graph} further proposes a simple yet effective graph-augmentation-free CL method for recommendation, which has distinct advantages in terms of both accuracy and training efficiency. Moreover, there are also a wave of methods that leverage CL with various auxiliary data for recommendation, \eg session graph~\cite{xia2021self}, social graph~\cite{yu2021self} and group-level hyper-graph~\cite{zhang2021double}.

Unfortunately, these approaches commonly perform CL via node- and graph-level augmentations in random, and abundant relations (\ie structural relations among queries, knowledge between head and tail queries and interaction between queries and intentions) in search scenarios are naturally ignored in contrastive pairs construction.


\subsection{Long-tail Issue}
Many real-world datasets encounter the long-tail issue, which severely threatens the task performance in various fields, \eg text classification~\cite{zhu2022enhanced}, visual recognition~\cite{zhang2021distribution} and recommendation~\cite{liu2020long,zhang2021model}.  A commonly adopted way to deal with the long-tail distribution follows the dataset-rebalancing strategies, including re-sampling data~\cite{he2009learning,buda2018systematic,chawla2002smote}, re-weighting loss~\cite{cao2019learning,cui2019class} and knowledge transferring from head categories~\cite{wang2017learning,tan2020equalization}.
When it comes to the area of recommendation and search, due to the popularity bias, the distribution of user feedback is very skewed, which heavily hinders the capability of recommendation models for tail items. To take advantage of learning tail items to widen user interests, more attention is shifted to tackling the long-tail issue. Typical re-weighting idea is naturally employed~\cite{yin2012challenging,park2008long}, while most of efforts treat it as a cold-start problem (\ie ``cold'' users and ``cold'' items). With the increasing availability of various kinds of auxiliary in online services, context information of ``cold'' users and items are greatly enriched through e bipartite graphs~\cite{he2020lightgcn}, social graphs~\cite{fan2019graph}, heterogeneous graphs~\cite{hu2018leveraging,shi2018heterogeneous} and knowledge graphs~\cite{wang2019kgat,feng2020atbrg}, which are effectively characterized with graph learning in the Euclidean space and the recently emerging hyperbolic space~\cite{chami2019hyperbolic,vinh2020hyperml,sun2021hgcf,du2022hakg}. Several evidences have showed that hyperbolic graph learning has a stronger ability to accommodate graphs with long-tail distributions~\cite{li2021hyperbolic,pan2021hyperbolic}.
On the other hand, inspired by the great success of self-supervised learning, numerous works argue that representations of long-tail users/items could be greatly enhanced with auxiliary contrastive supervisions~\cite{wu2021self,wei2021contrastive}.
It is also worth mentioning that the long-tail distribution in recommendation and search could be also alleviated through incorporating other learning methods, \eg meta-learning~\cite{zhang2021model,dong2020mamo} and active learning~\cite{zhu2019addressing}.

Different from these methods that purely enhance representations of tail items through side information and self-supervised learning, we focus on bridging head and tail queries with the interaction graph and hierarchical intention tree, such that knowledge from head queries containing rich  information could be effectively transferred to help the learning of tail queries.


\begin{figure*}
\centering
  \includegraphics[width=1.0\textwidth]{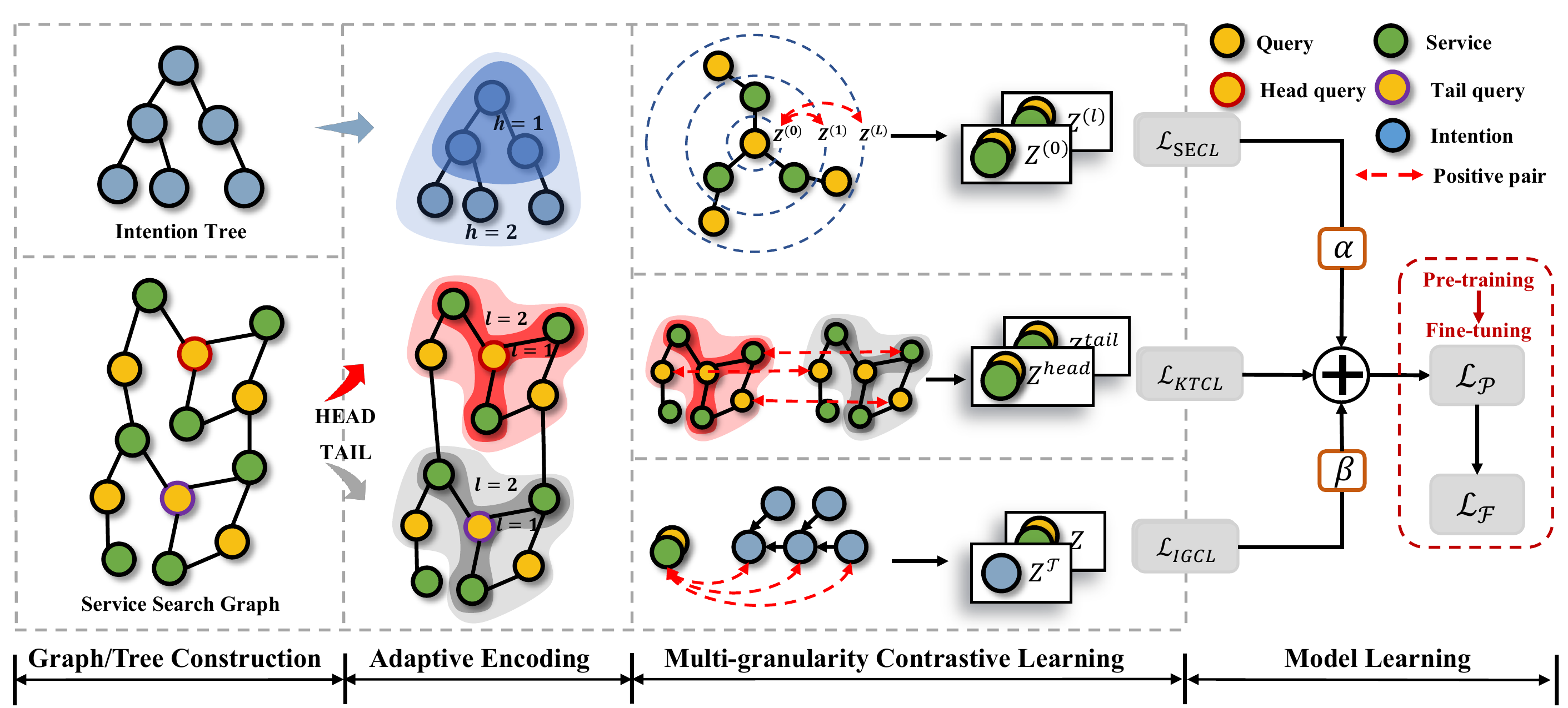}
  \caption{The overall architecture of the proposed {\model}. Best viewed in color.}
  \label{fig:framework}
\end{figure*}


%% file: sec-pre.tex
\section{Preliminaries \label{sec:pre}}

In this paper, we consider the long-tail issue in service search scenarios, where customer experience will be severely harmed if irrelevant services are provided.
With $n$ queries  $\mathcal{Q}=\{q_1,q_2,...,q_n\}$ and $m$ services $\mathcal{S}=\{s_1,s_2,...,s_m\}$, we define the user feedback as $\mathcal{Y} = \{(q, s, y_{q,s})\}$, each entry of which indicates whether a service $s$ is clicked by the user under the query $q$ (\ie the ground truth $y_{q,s} \in \{0, 1\}$).  
Due to the  popularity bias, the distribution of user feedback is very skewed, which heavily hinders the capability of search models for tail queries. 
To power the representation learning on tail queries, we seek to incorporate recently emerging graph learning, such that i) the context of tail query could be greatly enriched, 
and ii) knowledge from head queries containing rich information
could be effectively transferred to help the learning of tail
queries through this bridge.

\paratitle{Service Search Graph.} To achieve this goal, we frame our task in the setting of service search graph, which is  defined as $\mathcal{G} = \{\mathcal{V}, \mathcal{E}\}$, where $\mathcal{V} = \{\mathcal{Q} \bigcup \mathcal{S}$\} denotes the node set and   $\mathcal{E} = \{(q,s)| q \in \mathcal{Q}, s \in \mathcal{S}\}$ denotes the edge set. Moreover, each node $v$ is also associated with several attributes, denoted as $x_v$. In particular, with the consideration of the characteristics of the service search scenarios, we carefully establish the connection between a query $q$ and a service $s$ with the following conditions to avoid underline noise:
\begin{itemize}[leftmargin=*]
    \item \textbf{Interaction condition}: the service $s$ is clicked by users with the query $q$ in the past 30 days. And the click-through rate (\ie CTR) is also reserved as the edge feature for the following graph learning.
    \item \textbf{Correlation condition}: the target query $q$ and service $s$ are expected to share the same correlation, \eg city, brand and category. Similarly, this information would form the edge feature to guide the model learning.
\end{itemize}
Here, we organize the scenario of service search as a well-established graph that can comprehensively integrate rich features and interaction information. As a promising direction, graph neural networks (GNNs) have shown its potential in structure feature extraction and has been successfully applied into recommendation and search.

\paratitle{Recap of GNNs in Recommendation.}
Recently, GNNs are widely adopted in recommender systems~\cite{humerit}, which follow the information aggregation scheme to obtain informative representation for users and items~\cite{he2020lightgcn,wang2019kgat}. Through stacking multiple GNN layers, high-order connections could be fully summarized. Generally, such a procedure of a $L$-layers GNN could be formulated as following three stages (take the users' side as an example):
\begin{equation}
    \begin{split}
        \bm{m}^{(l)}_u & = f_{aggregate}(\{\bm{z}^{(l)}_{u'} | u' \in \mathcal{N}_u\}; \Theta^{\mathcal{A}}), \\
        \bm{z}^{(l+1)}_u & = f_{update}(\bm{z}^{(l)}_u, \bm{m}^{(l)}_u; \Theta^{\mathcal{U}}), \\
        \bm{z}_u  & =  f_{readout}(\{\bm{z}^{(0)}_u, \bm{z}^{(1)}_u, \cdots, \bm{z}^{(L)}_u\}; \Theta^{\mathcal{R}}),
    \end{split}
\end{equation}
where $\mathcal{N}_u$ represents the neighbor set of the user $u$. Here, the GNN initializes the $\bm{z}^{(0)}_u$ with the original attributes or learnable embedding table. In each iteration, it aggregates $(l)^{th}$ layer’s representations of neighbors (\ie $\bm{z}^{(l)}_{u'}$) via $f_{aggregate}(\cdot; \Theta^{\mathcal{A}})$, and then generates the $(l+1)^{th}$ layer’s representation $\bm{z}^{(l+1)}_u$ via $f_{update}(\cdot; \Theta^{\mathcal{U}})$ with aggregated representation (\ie $\bm{m}^{(l)}_u$). After $L$-layers recursive propagation, the final representation $\bm{z}_u$ is produced through  $f_{readout}(\cdot; \Theta^{\mathcal{R}})$.

Our approach borrows the idea of GNNs for enriching tail queries as well as learning discriminative representations in service search scenarios via subtly designing the above-mentioned three core stages.

\paratitle{Intention Tree.}
Distinct from recommender systems, users' search behaviors for services are more intention oriented. Hence, in our scenarios, each query or service is attached with at most $5$-level intentions. Naturally, there are various relations and hierarchical dependencies among intentions, forming a tree.

\begin{definition}{\textbf{Intention Tree}}
A intention tree is defined as $\mathcal{T}  = \{\mathcal{V}^{\mathcal{T}}, \mathcal{E}^{\mathcal{T}}\}$. In particular, the node set $\mathcal{V}^{\mathcal{T}}$ consists of non-leaf and leaf nodes, where each node except the root node has one parent and an arbitrary number of children. An edge $e^t = (v^t_i, v^t_j) \in \mathcal{E}^{\mathcal{T}}$ indicates that $v^t_i$ is the parent of $v^t_j$, commonly containing more coarse-grained concepts.
\end{definition}


An example of intention trees is illustrated in the Fig.~\ref{fig:example}, in which
each circle represents an intention. Note a variety of intention trees exist in our service search scenario, which could be regarded as ``forest''.
In this paper, we adopt the intention tree as a kind of general knowledge to supplement the queries/services modeling with intention-level alignment and  hierarchical structure characterization.

%% file: tab/dataset.tex
\begin{table}
	\caption{The statistics of six datasets. Search PVs are not presented in public datastes due to the unavailability of the exposure information.}
	\label{tab:dataset}
\centering
\setlength{\tabcolsep}{2.5mm}{
	\begin{tabular}{c|c|c|c|c}
		\toprule
        \multicolumn{2}{c|}{Industrial datasets} & {Sep. A} & {Sep. B} & {Sep. C} \\
        \midrule
         \multirow{2}{*}{\# Queries} &{Head} & {1.18\%} & {1.64\%} & {1.51\%} \\
        {} &{Tail} & {98.82\%} & {98.36\%} & {98.49\%} \\
        \midrule
        \multirow{2}{*}{\# Search PV} &{Head} & {93.57\%} & {94.07\%} & {93.82\%} \\
        {} &{Tail} & {6.43\%} & {5.93\%} & {6.18\%} \\
        \midrule
        \multicolumn{2}{c|}{\# Train} & {$1.39 \times 10^9$} & {$1.25 \times 10^9$} & {$1.24 \times 10^9$} \\
        \midrule
        \multicolumn{2}{c|}{\# Validation} & \multicolumn{3}{c}{$1 \times 10^6$} \\
        \midrule
        \multicolumn{2}{c|}{\# Test} & \multicolumn{3}{c}{$1 \times 10^6$} \\
        \midrule
        \midrule
        \multicolumn{2}{c|}{Public datasets} & {Software} & { Video game} & { Music} \\
        \midrule
         \multirow{2}{*}{\# Queries} &{Head} & {10.95\%} & {3.62\%} & {3.63\%} \\
        {} &{Tail} & {89.05\%} & {96.38\%} & {96.37\%} \\
        \midrule 
        \multicolumn{2}{c|}{\# Train} & {$8.08 \times 10^3$} & {$3.13 \times 10^5$} & {$1.46 \times 10^5$} \\
        \midrule 
        \multicolumn{2}{c|}{\# Validation} & {$8.86 \times 10^2$} & {$3.49 \times 10^4$} & {$1.61 \times 10^4$} \\
        \midrule
        \multicolumn{2}{c|}{\# Test} & {$3.84 \times 10^3$} & {$1.49 \times 10^5$} & {$6.94 \times 10^4$} \\
        
		\bottomrule
	\end{tabular}}
\end{table}

%% file: tab/dataset_graph.tex
\begin{table}
	\caption{The statistics of search service graph and intention tree.}
	\label{tab:dataset-graph}
\centering
\setlength{\tabcolsep}{2.0mm}{
	\begin{tabular}{c|c|c|c|c}
		\toprule
        \multicolumn{2}{c|}{} & {Head } & {Tail} &  \multirow{2}{*}{Intention tree}\\
		\multicolumn{2}{c|}{} & \multicolumn{2}{c|}{Service search graph} & {} \\
		
        \midrule
        \multirow{2}{*}{Sep. A, B, C}&{\# Nodes} & {$1.75 \times 10^6$} & {$2.40 \times 10^6$} & {$3.25 \times 10^3$} \\
        \cmidrule{2-5}
        {}&{\# Edges} & {$3.75 \times 10^5$} & {$2.00 \times 10^6$} & {$3.28 \times 10^3$} \\
        \midrule
        \midrule
        \multirow{2}{*}{Software}&{\# Nodes} & {$8.61 \times 10^2$} & {$2.42 \times 10^3$} & {$4.63 \times 10^3$} \\
        \cmidrule{2-5}
        {}&{\# Edges} & {$2.79 \times 10^3$} & {$9.10 \times 10^3$} & {$5.10 \times 10^3$} \\
        \midrule
        \midrule
        \multirow{2}{*}{Video game}&{\# Nodes} & {$1.65 \times 10^4$} & {$7.06 \times 10^4$} & {$2.88 \times 10^4$} \\
        \cmidrule{2-5}
        {}&{\# Edges} & {$8.16 \times 10^4$} & {$3.92 \times 10^5$} & {$3.01 \times 10^4$} \\
        \midrule
        \midrule
        \multirow{2}{*}{Music}&{\# Nodes} & {$8.79 \times 10^3$} & {$3.71 \times 10^4$} & {$8.42 \times 10^2$} \\
        \cmidrule{2-5}
        {}&{\# Edges} & {$2.95 \times 10^4$} & {$1.90 \times 10^5$} & {$1.12 \times 10^3$} \\
		\bottomrule
	\end{tabular}}
\end{table}

%% file: tab/eff_exp_table.tex
\begin{table*}
	\caption{Performance comparison \wrt AUC with baselines on three industrial datasets and three public datasets.}
	\label{tab:baseline}
\centering
\setlength{\tabcolsep}{3.0mm}{
	\begin{tabular}{c|c|c|c|c|c|c|c|c|c}
		\toprule
        \multirow{2}{*}{Industrial datasets} & \multicolumn{3}{c|}{Sep. A} & \multicolumn{3}{c|}{Sep. B} & \multicolumn{3}{c}{Sep. C}\\
        \cmidrule{2-10}
        {} & {Head} & {Tail} & {Overall}  & {Head} & {Tail} & {Overall}  & {Head} & {Tail} & {Overall}\\
        \midrule
        {Wide\&Deep} & {0.8676} & {0.7581} & {0.8629} & {0.8627} & {0.7484} & {0.8518} & {0.8618} & {0.7468} & {0.7352} \\
        {LightGCN} & {0.9199} & {0.8205} & {0.9156} & {0.9084} & {0.8172} & {0.9034} & {0.9060} & {0.8171} & {0.9012} \\
        {KGAT} & {0.9310} & {0.8169} & {0.9264} & {0.9290} & {0.8245} & {0.9239} & {0.9239} & {0.8191} & {0.9189} \\
        {SGL} & {0.9291} & {0.8024} & {0.9241} & {0.9312} & {0.8175} & {0.9258} & {0.9227} & {0.8064} & {0.9173} \\
        {SimSGL} & {0.9333} & {0.8142} & {0.9286} & {0.9140} & {0.7781} & {0.9073} & {0.9202} & {0.8088} & {0.9149} \\
        \midrule
        {{\model}} & {\textbf{0.9361}} & {\textbf{0.8285}} & {\textbf{0.9320}} & {\textbf{0.9332}} & {\textbf{0.8374}} & {\textbf{0.9284}} & {\textbf{0.9248}} & {\textbf{0.8310}} & {\textbf{0.9203}}\\
        {(\emph{v.s.} best)} & {\textbf{(+0.65\%)}} & {\textbf{(+2.50\%)}} & {\textbf{(+0.78\%)}} & {\textbf{(+0.46\%)}} & {\textbf{(+3.96\%)}} & {\textbf{(+0.61\%)}} & {\textbf{(+0.22\%)}} & {\textbf{(+3.73\%)}} & {\textbf{(+0.32\%)}}\\
        \midrule
        \multirow{2}{*}{Public datasets}  & \multicolumn{3}{c|}{Software} & \multicolumn{3}{c|}{Video game} & \multicolumn{3}{c}{Music} \\
        \cmidrule{2-10}
        {} & {Head} & {Tail} & {Overall} & {Head} & {Tail} & {Overall} & {Head} & {Tail} & {Overall}  \\
        \midrule
        {Wide\&Deep} & {0.7513} & {0.7483} & {0.7483} & {0.6443} & {0.5564} & {0.5712} & {0.6074} & {0.5841} & {0.5972} \\
        {LightGCN} & {0.8481} & {0.8202} & {0.8263} & {0.8025} & {0.7475} & {0.7562} & {0.6585} & {0.6635} & {0.6610}\\
        {KGAT} & {0.8190} & {0.8342} & {0.8280} & {0.7931} & {0.7605} & {0.7637} & {0.6750} & {0.6813} & {0.6746}\\
        {SGL} & {0.8182} & {0.8099} & {0.7989} & {0.7878} & {0.7610} & {0.7654} & {0.6512} & {0.6712} & {0.6515}\\
        {SimSGL} & {0.8571} & {0.8144} & {0.8248} & \textbf{0.8075} & {0.7671} & {0.7746} & {0.6512} & {0.6712} & {0.6515}\\
        \midrule
        {{\model}} & {\textbf{0.8605}} & {\textbf{0.8399}} & {\textbf{0.8436}} & {0.8060} & {\textbf{0.7713}} & {\textbf{0.7763}} & \textbf{0.6758} & {\textbf{0.7094}} & {\textbf{0.6906}} \\
        {(\emph{v.s.} best)} & {\textbf{+0.95\%}} & {\textbf{+1.71\%}} & {\textbf{+4.76\%}} & {-0.48\%} & {\textbf{+1.57\%}} & {\textbf{+0.62\%}} & \textbf{+0.45\%} & {\textbf{+15.50\%}} & {\textbf{+9.16\%}} \\
		\bottomrule
	\end{tabular}}
\end{table*}

%% file: tab/eff_exp_table_2.tex
\begin{table}
	\caption{Performance comparison \wrt GAUC and NDCG@10 for tail queries on industrial datasets. We also report the improvement ratio  of all methods over the  LightGCN.} 
	\label{tab:baseline_2}
\centering
\setlength{\tabcolsep}{3.0mm}{
	\begin{tabular}{c|c|c|c}
		\toprule
        {} & {} &  {GAUC} & {NDCG@10}\\
		\midrule
		{} & {Wide\&Deep} & {0.6513 (-22.41\%)} & 	{0.8113 (-3.48\%)}   \\
		{} & {LightGCN} & {0.6950 (-) } &	{0.8406 (-)}  \\
		{Sep. A} & {KGAT} & {0.7083 (+6.82\%)} &	{0.8509 (+1.23\%)}   \\
		{} & {SGL} & {0.7067 (+6.00\%)} & {0.8412 (+0.07\%)}   \\
		{} & {SimSGL} & {0.7078 (+6.56\%)} & {0.8463 (+0.67\%)}   \\
		{} & {\model} & {\textbf{0.7103 (+7.84\%)}} & {\textbf{0.8596 (+2.26\%)}}   \\
		\midrule
		{} & {Wide\&Deep} & 0.6574 (-15.37\%) &	0.8082 (-2.86\%)  \\
        {} & {LightGCN} & 0.6860  (-)  &	0.8320 (-)  \\
        {Sep. B} & {KGAT} & 0.7085  (+12.09\%)  &	0.8469 (+1.79\%)  \\
        {} & {SGL} & 0.7073  (+11.45\%)  &	0.8456  (+1.63\%) \\
        {} & {SimSGL} & 0.6944  (+4.51\%)  &	0.8271 (-0.59\%)  \\
        {} & {\model} & \textbf{0.7119  (+13.92\%)}  &	\textbf{0.8591 (+3.26\%)}  \\
        \midrule
        {} & {Wide\&Deep} & 0.6522 (-17.82\%) &	0.8107 (-3.23\%)  \\
        {} & {LightGCN} & 0.6852 (-) &	0.8386 (-)  \\
        {Sep. C} & {KGAT} & 0.7014 (+8.75\%) &	0.8500 (+1.36\%) \\
        {} & {SGL} & 0.7003 (+8.15\%)	& 0.8409 (+0.27\%)   \\
        {} & {SimSGL} & 0.6954 (+5.51\%) &	0.8436 (+0.59\%) \\
        {} & {\model} & \textbf{0.7056 (+11.02\%)} &	\textbf{0.8593 (+2.47\%)} \\
		\bottomrule
	\end{tabular}}
\end{table}

%% file: sec-con.tex
\section{Conclusion \label{sec:con}}
In this paper, we proposed a novel framework called {\model} tailored for the long-tail issue in service search scenarios. 
First of all, we elaborately constructed a service search graph based on well-designed interaction and correlation conditions.
On top of the well-established service search graph and intention trees, we employed an adaptive encoder to obtain informative representations for queries and services as well as hierarchical structure aware representation for intention. 
Next, we equipped GARCIA with a novel multi-granularity contrastive learning module, which powers representations through knowledge transfer, structure enhancement and intention generalization,
followed by a pre-training \& fine-tuning based learning schema.
Extensive experiments in both offline and online environments indicated the superior
performance of GARCIA.

As for future works, we will investigate into how to split queries into multiple groups via frequency in an adaptive manner and perform effective knowledge transfer between query groups with different frequencies. Moreover, incorporating semantic-level information through text mining modules (\eg BERT~\cite{devlin2018bert}) into {\model} is also an interesting direction.